\documentclass[twoside]{article}
\usepackage{ecj,palatino,epsfig,latexsym,natbib}
\usepackage{booktabs}
\usepackage{longtable}
\usepackage{xcolor}
\usepackage{subcaption}
\usepackage{amsmath}
\usepackage{hyperref}
\usepackage{cleveref}

\newcommand{\F}{\textbf{\texttt{F}}}
\newcommand{\X}{\textbf{\texttt{X}}}
\newcommand{\V}{\textbf{\texttt{V}}}
\newcommand{\N}{\textbf{\texttt{N}}}
\newcommand{\D}{\textbf{\texttt{D}}}
\newcommand{\I}{\textbf{\texttt{I}}}

\begin{document}

\ecjHeader{x}{x}{xxx-xxx}{201X}{PyXla Toolbox for Explainable Landscape Analysis}{T. Ombaso et al.}
\title{\bf A Flexible and Generic Approach for Explainable Landscape Analysis\\and the \texttt{pyXla} Toolbox}

\author{\textbf{Tony Ombaso} \\
        u25700155@tuks.co.za \\
        Department of Computer Science, University of Pretoria, Pretoria, South Africa
\AND
        \textbf{Anna S. Bosman} \\
        anna.bosman@up.ac.za \\
        Department of Computer Science, University of Pretoria, Pretoria, South Africa \\
        African Institute for Data Science and Artificial Intelligence (AfriDSAI), University of Pretoria, Pretoria, South Africa
\AND
        \textbf{Arnaud Liefooghe} \\
        arnaud.liefooghe@univ-littoral.fr \\
        Univ. Littoral C\^ote d'Opale, UR 4491, LISIC, F-62100 Calais, France
\AND
        \textbf{Katherine M. Malan} \\
        malankm@unisa.ac.za \\
        Department of Decision Sciences, University of South Africa, Pretoria, South Africa
\AND
        \textbf{S\'ebastien Verel} \\
        verel@univ-littoral.fr \\
        Univ. Littoral C\^ote d'Opale, UR 4491, LISIC, F-62100 Calais, France}
\maketitle

\begin{abstract}
Landscape analysis has been successfully applied to understand complex optimisation problems, gain insights into algorithm behaviour, and automate algorithm selection and configuration. Although many landscape analysis techniques have been developed over the last decades, it remains difficult for researchers and practitioners to decide which approaches are appropriate and to implement them in practice. Some tools are available, but these are either restricted to particular problem domains (e.g., unconstrained black-box continuous optimisation), or are limited in what they model and measure. In addition, output from landscape analysis is often not easily interpretable, especially when computed landscape features do not correspond with aspects of problems that practitioners are familiar with. In this paper, we introduce a principled approach for \emph{explainable landscape analysis} (XLA) with an associated Python package called \texttt{pyXla}. The approach is generic in that it applies to problems with different representations (continuous or combinatorial), with single or multiple objectives, with or without constraints. The extent of analysis provided by the XLA framework depends on the data available, with richer analysis offered as additional information is provided by the user. We demonstrate the explainable output produced by \texttt{pyXla} on a selection of hand-crafted problems with diverse landscape characteristics.
\end{abstract}

\begin{keywords}
Randomised optimisation, 
continuous and combinatorial domains,
single and multiple objectives,
unconstrained and constrained problems,
landscape analysis,
explainability,
\texttt{pyXla} toolbox.
\end{keywords}



\section{Introduction}
When practitioners face their own optimisation problem, they often lack clear guidance on how to choose, configure, or design an algorithm to solve it.
%
Landscape analysis (LA) aims to understand optimisation problems by casting the search space into a topology, referred to as a fitness landscape, where fitness reflects the quality of solutions~\citep{malan2013survey}.
It helps reveal what makes a problem hard and translates those insights into actionable guidance for the problem at hand.
This is particularly relevant in black-box optimisation, where only limited information about the problem is available.
A landscape comprises three components: (1) a set of candidate solutions to the optimisation problem, (2) a notion of distance between solutions or a neighbourhood that connects them, and (3) an objective (fitness) function that maps each solution to its corresponding fitness value~\citep{stadler2002fitness,merz2004}.
LA is concerned with developing techniques to analyse optimisation problems by characterising landscape features. Features are quantitative or visual representations of the geometric, topological, and statistical properties of the landscape~\citep{gallagher2024towards}.
Early LA features include the number and distribution of local optima~\citep{kauffman1993}, the landscape ruggedness~\citep{weinberger1991}, and the fitness distance correlation~\citep{jones1995fitness}. These have evolved into more advanced representations such as local optima networks~\citep{ochoa2014}.

Today, exploratory landscape analysis (ELA)~\citep{mersmann2011exploratory} is one of the dominant streams of LA. It uses large numbers of low-level features that can be extracted cheaply to characterise unknown optimisation problems.
ELA features include measures of convexity, statistical distribution of objective values, level-set features, meta-model features, local search features, curvature features, and so on.
However, some of these low-level features are not directly interpretable~\citep{kerschke2019comprehensive}.
Hence,~\citet{mersmann2011exploratory} propose relating them to high-level properties such as separability, global structure, multimodality, and search space homogeneity.
The original aim of ELA was \emph{automated} problem analysis, where many low-level features are used to train machine learning models to distinguish between problems for automated algorithm selection~\citep{kerschke2019automated,seiler2025deep}.
However, applying machine learning to uninterpretable low-level features entails a loss of explainability: the focus has shifted from building an intuitive understanding of the nature of optimisation problems and algorithm performance to relying on black-box problem characterisation and algorithm selection~\citep{renau2021towards,gallagher2024towards}.
We further observe that, although many (E)LA features do carry meaning, that meaning has, to some extent, been diluted over time. This is partly because ELA analyses typically consider a high-dimensional feature space and tend to favour machine learning pipelines. As a result, it has become difficult to state concretely what characterises different problem classes in terms of ELA features; for example, which precise features distinguish problems with high levels of neutrality from those with low neutrality.

Despite the rapid growth of landscape features for optimisation, much of this recent progress has been driven by automation and predictive performance models, rather than by metrics that help practitioners understand why an algorithm behaves as it does on a given problem. In its early years, explainability and interpretability were central goals of LA~\citep{stadler2002fitness,merz2004}. We argue for a return to that perspective: explainability is a key aim of LA and an important property, especially in the context of explainable artificial intelligence. Our key premise is that a landscape feature is most valuable when it provides actionable meaning for the practitioner, not necessarily when it achieves high predictive accuracy. This matters because predictive performance is inherently context-dependent, for example, when forecasting the performance of a specific algorithm or selecting among a given portfolio of algorithms within a particular problem class.

Over the years, many LA techniques have been developed.
However, the choice of which techniques to use and their implementation is non-trivial, more so for those who are not experienced in the field. 
To address this issue, software libraries aggregating LA techniques, such as \texttt{flacco}, 
have been developed. The \texttt{flacco} library conveniently provides more than 300 LA features for the R language ecosystem~\citep{kerschke2019comprehensive}. Developed more recently, the \texttt{pflacco} library provides the same convenience to the Python community whilst adding a few newer LA techniques~\citep{prager2024pflacco}.
However, the suitability of (\texttt{p})\texttt{flacco} for landscape understanding is limited because of its focus on ELA techniques, and specialisation to continuous problems (with box constraints only). 
Since the majority of ELA features lack intuitive interpretability, the low-level features extracted by the implemented techniques are more suited for automated algorithm analysis rather than problem understanding and visualisation~\citep{kerschke2019comprehensive}.
Likewise, \texttt{Paradiseo} provides a comprehensive metaheuristics framework~\citep{paradiseo}, including LA features for (mostly unconstrained, combinatorial) single-objective optimisation~\citep{paradiseo_mo}. While this landscape analysis functionality exists, the \texttt{Paradiseo} framework and software is not actively maintained. In addition, its focus is more on algorithm design than on problem understanding and visualisation.
To the best of the authors' knowledge, no unified library combines intuitively interpretable LA features while covering all problem domains (single- and multi-objective, constrained and unconstrained, combinatorial and continuous).

In this paper, we propose a new framework for \emph{explainable landscape analysis} (XLA) that is both flexible and generic.
We refer to this approach as XLA to emphasize its focus on explainability and to distinguish it from black-box LA techniques that are currently proliferating in the field and do not account for the semantic meaning of LA features.
Explainability is achieved by incorporating LA features that are meant for intuitive interpretation, and are accompanied by visualisations alongside metrics.
Flexibility and generality are achieved by separating data loading and sampling from the actual analysis, such that all the techniques implemented operate on a unified input format.
As a result, the proposed framework can produce results even with a minimum of input components; for instance, a sample of objective values is sufficient to produce meaningful output. As more information is provided, such as violation values, neighbourhood information, or solution samples, additional features become applicable.
The inclusion of sampling algorithms in the framework further contributes to this flexibility by allowing the generation of inputs from a specified domain.
The property of being generic refers to allowing analysis across different problem domains: combinatorial and continuous, constrained and unconstrained, single-objective and multi-objective.

We implement the proposed XLA framework as an open-source Python package called \texttt{pyXla} (\textbf{Py}thon package for e\textbf{X}plainable \textbf{l}andscape \textbf{a}nalysis), made available at \href{https://pypi.org/project/pyxla}{https://pypi.org/project/pyxla}. We validate \texttt{pyXla} on a selection of continuous and discrete optimisation problems with and without constraints or multiple objectives, and demonstrate the interpretable visualisations and metrics that \texttt{pyXla} can provide. 

The rest of the paper is structured as follows. Section~\ref{sec:overview} provides a high-level overview of the proposed XLA framework. Section~\ref{sec:pyxla} presents XLA features and discusses the \texttt{pyXla} implementation. Section~\ref{sec:examples} provides illustrative examples, demonstrating \texttt{pyXla} in action. Section~\ref{sec:conclusion} concludes the paper and discusses further research.%





\section{Overview of the Proposed XLA Framework}\label{sec:overview}

\begin{figure}[t!]
   \centering
   \includegraphics[width=\textwidth]{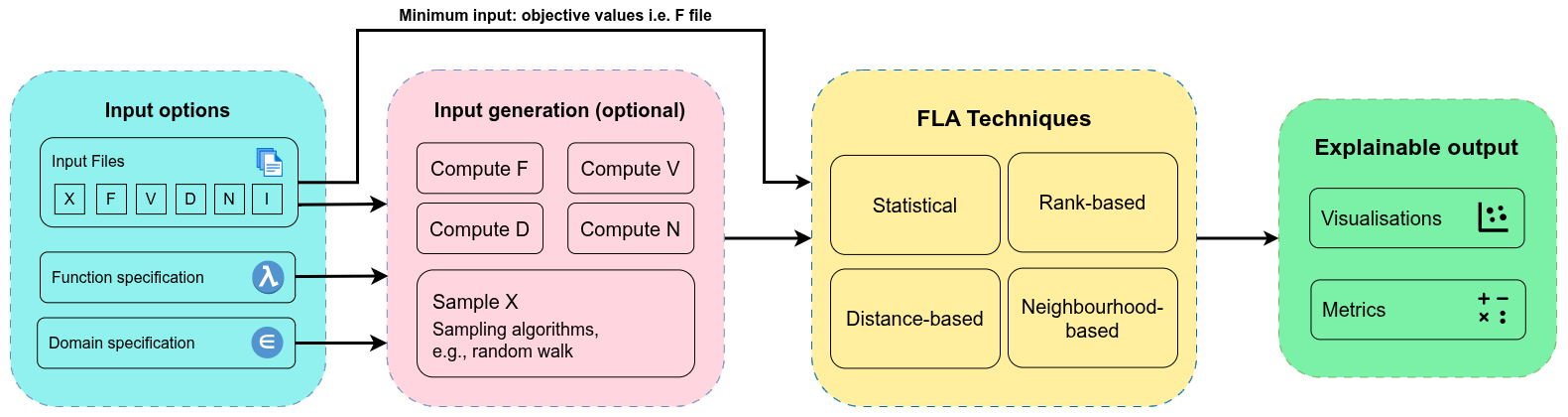}
   \caption{A high-level diagram of the proposed XLA framework. Data loading and sampling are separated from analysis.}
   \label{fig:high-level}
\end{figure}

The proposed framework comprises two key components that are separate from one another: data representation and analysis. 
Data representation entails data loading and sampling. 
Analysis involves the application of LA techniques to the loaded or sampled data.
\Cref{fig:high-level} is a high-level diagram of the proposed framework.

Data loading and sampling utilise an input format that specifies file names for different kinds of data, as follows:
\begin{itemize}
\item sampled solutions are identified by the letter \X,
\item objective values by the letter \F,
\item constraint violation values correspond to \V,
\item distance information to \D,
\item neighbourhood information to \N, and
\item other additional data to \I.
\end{itemize}
As such, XLA features are all relative to the provided sample (or a full enumeration, if available). However, the user retains full control over the budget allocated to their sample (i.e., XLA performs no additional evaluations).
If data is not already sampled or generated, the framework supports sampling and input generation if needed, where data points can be sampled from a domain, or various input files can be generated from existing ones. To support sampling, the framework provides several algorithms, including random uniform sampling, Latin hypercube sampling, random walk sampling~\citep{weinberger1990}, adaptive walk sampling~\citep{kauffmanGeneralTheoryAdaptive1987}, Hilbert curve sampling~\citep{pienaarHilbertCurvesEfficient2024}, and even full enumeration for small combinatorial solution spaces.

The analysis component comprises LA techniques that are categorised into statistical, rank-based, distance-based, and neighbourhood-based techniques. All included LA techniques produce both visual and numerical outputs for explainability purposes.
These LA techniques are implemented in a generic way where possible, such that the same techniques can be used to analyse not only fitness values but also violation values, in the case of constrained problems. For instance, the fitness distance correlation technique~\citep{jones1995fitness} is extended to violation values, yielding violation distance correlation, thereby enabling analysis of violation landscapes.

\section{XLA Features and the \texttt{pyXla} Package}\label{sec:pyxla}
The \texttt{pyXla} package is the implementation of the proposed XLA framework in the Python programming language. This section discusses various components of \texttt{pyXla}. \Cref{subsec:input} describes how the package accepts input 
and highlights the auxiliary functions included in the package to support data generation. \Cref{subsec:features} outlines the LA techniques currently provided with the package.

\begin{table}[!t]
\centering
\caption{Description of the input data in \texttt{pyXla}. The \texttt{id} in \X, \F , \V, \I, and \texttt{id1} and \texttt{id2} in \D\ and \N\ link the data between files.}%
\label{tab:inputData}%
\begin{tabular}{c|p{11.65cm}}
\toprule
\textbf{Input}   &   \textbf{Description} \\
\midrule
\X  &   \textbf{Solutions in the sample}\\
    &   Example headings for five decision variables: \texttt{id x1 x2 x3 x4 x5}\\
    &  If not available as input data, \X\ can be generated by \texttt{pyXla} using: specification of the domain of each variable with sampling strategy.\\
\midrule
\F  &   \textbf{Objective values of the sample of solutions}\\
    &   Example headings for three objectives: \texttt{id f1 f2 f3}\\
    &   If not available as input data, \F\ can be generated by \texttt{pyXla} using: specification of objective function(s) and \X.\\
\midrule
\V  & \textbf{Violation extent of each constraint for each solution in the sample}\\
    & Example headings for two constraints: \texttt{id v1 v2}\\
    & If not available as input data, \V\ can be generated by \texttt{pyXla} using: specification of violation function(s) and \X.\\
\midrule
\D  & \textbf{Pairwise distance between solutions}\\
    & Headings: \texttt{id1 id2 d} \\
    & If not available as input data, \D\ can be generated by \texttt{pyXla} using: specification of a distance metric function between pairs of solutions and \X. \\
\midrule
\N  & \textbf{Neighbouring pairs of solutions}\\
    & Headings: \texttt{id1 id2} \\
    & If not available as input data, \N\ can be generated by \texttt{pyXla} using: specification of a neighbourhood function that tests if two solutions are neighbours, or \D\ with a specified $\epsilon$ for defining the neighbourhood size or $k$ for the number of closest solutions defining the neighbourhood.\\
\midrule
\I  & \textbf{Additional information}\\
     &   Example headings for two additional information features: \texttt{id i1 i2}\\
    & This file is used to specify any other additional information not captured by any other components above. For instance, in the context of neural networks, gradient information can be provided in this file.\\
\bottomrule
\end{tabular}
\end{table}

\subsection{Input Data}\label{subsec:input}
Table~\ref{tab:inputData} describes the input used by \texttt{pyXla} to conduct analysis. Only the objective (fitness) function values (\F) are required to conduct basic analysis. As additional inputs are provided, richer analysis can be conducted. It is a requirement that all of the inputs are with respect to the same sample of solutions, and each solution in the sample has a unique identifier used across the input files.

The user has the option of either providing the input in the form of text (CSV) files as specified, or of providing alternative information, in which case \texttt{pyXla} will conduct appropriate computation to generate the input required for analysis. The alternative information needed is specified for each input in Table~\ref{tab:inputData}.

\subsubsection*{Auxiliary Functions for Input Data Generation}\label{subsec:aux-functions}
In the absence of sampled data, \texttt{pyXla} supports the generation of input data based on the specification of a domain or sampling function.
\texttt{pyXla} provides auxiliary functions that allow the generation of any type of input file supported. The auxiliary functions are presented in~\Cref{tab:supportingFunctions}. All auxiliary functions, excluding the \texttt{sample\_X} function, require that an \X~file be present. If no \X~file is provided, it can be generated by sampling the problem space using the \texttt{sample\_X} function.

\begin{table}[!t]
\centering
\caption{Description of the auxiliary functions in \texttt{pyXla}.}
\label{tab:supportingFunctions}
\begin{tabular}{c|p{10.7cm}}
\toprule
\textbf{Function}   &   \textbf{Description} \\
\midrule
\texttt{sample\_X}  &  Takes a domain specification for each variable with a sampling strategy as input and generates an \X~file.\\
\midrule
\texttt{compute\_F}  &   Takes an objective function as input and generates an \F~file. The \X~file is required.\\
\midrule
\texttt{compute\_V}  & Takes a violation function as input and generates a \V~file. The \X~file is required.\\
\midrule
\texttt{compute\_D}  & Takes a distance metric function as input and generates a \D~file. The \X~file is required.\\
\midrule
\texttt{compute\_N}  & Takes an arbitrary neighbourhood function that determines if two solutions are neighbours and generates an \N~file. 
The \X~file is required.\\
\bottomrule
\end{tabular}
\end{table}

\begin{figure}[!t]
\centering
\begin{subfigure}{0.3\textwidth}
    \includegraphics{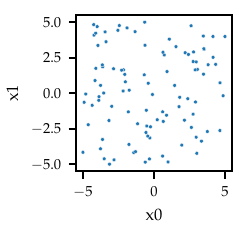}
    \caption{Uniform random}
    \label{fig:sample-random}
\end{subfigure}
\begin{subfigure}{0.3\textwidth}
    \includegraphics{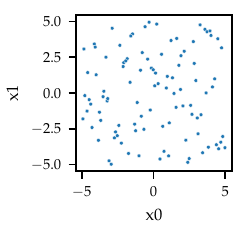}
    \caption{Latin hypercube}
    \label{fig:sample-lhs}
\end{subfigure}
\begin{subfigure}{0.3\textwidth}
    \includegraphics{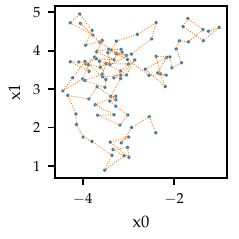}
    \caption{Random walk}
    \label{fig:sample-random-walk}
\end{subfigure}
\begin{subfigure}{0.3\textwidth}
    \includegraphics{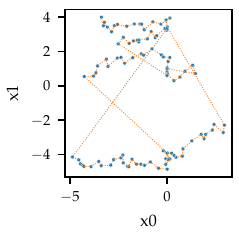}
    \caption{Adaptive walk}
    \label{fig:sample-adaptive-walk}
\end{subfigure}
\begin{subfigure}{0.3\textwidth}
    \includegraphics{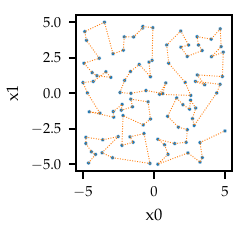}
    \caption{Hilbert curve}
    \label{fig:sample-hilbert-curve}
\end{subfigure}
\caption{Comparison of the different sampling algorithms in 2D for 100 samples. Dots represent the sampled points. For ordered sampling methods, the connections between them indicate the sampling order. All samples lie within the bounding box [-5, 5]. Random walk (\Cref{fig:sample-random-walk}) and adaptive walk sampling (\Cref{fig:sample-adaptive-walk}) are carried out with a step size of 0.5.
The adaptive walk, which is guided by the objective function, is based on the sphere function (\Cref{eq:sphere}) in \Cref{fig:sample-adaptive-walk}.}
\label{fig:sampling}
\end{figure}

The choice of the right sampling algorithm is critical for extracting meaningful information from the landscape. Because landscape features inherently depend on the sample, different sampling algorithms can yield different perspectives of the same problem instance.
Five sampling algorithms are implemented in \texttt{pyXla}: random uniform sampling, 
Latin hypercube sampling, 
random walk sampling, 
adaptive walk sampling, 
and Hilbert curve sampling. 
\Cref{fig:sampling} shows a comparison of the five sampling algorithms, run with the same seed to produce 2-dimensional samples of the same length within the bounding box~$[-5,5]$.

Random uniform sampling 
and Latin hypercube sampling (LHS) are implemented due to their simplicity and ubiquity in the field~\citep{renau2020exploratory}.
A walk is an ordered sequence of solutions in which each sample is a neighbour of the previous one.
During a random walk, a neighbour is chosen at random at each step; this is typically used to characterise the ruggedness of the landscape~\citep{weinberger1990,malanProgressiveRandomWalk2014}.
By contrast, an adaptive walk selects an improving neighbour at each step until it gets trapped into a local optimum~\citep{kauffmanGeneralTheoryAdaptive1987}.
%
This additional sampling strategy targets features suited to measuring hill-climbing algorithm traces, which is key to capturing multimodality (local optima and their basins of attraction).
In \texttt{pyXla}, the adaptive walk implementation is based on a simple hill climber that sequentially explores random neighbours and immediately accepts any that yield an improvement.
Finally, Hilbert curve sampling (HCS) is included to ensure adequate coverage of the search space, with the samples being spatially correlated when they are mapped to a 1-dimensional space~\citep{pienaarHilbertCurvesEfficient2024}. 
Spatial correlation eliminates the need to re-order the samples to determine neighbourhood, and therefore is more efficient than LHS, which may need post-factum ordering for some LA metrics. 
Another auxiliary function based on the Hilbert curve~\citep{pienaarHilbertCurvesEfficient2024} is also implemented to efficiently order unordered samples generated by other sampling algorithms such as the LHC.

\subsection{XLA Features}\label{subsec:features}
\begin{figure}[!htb]
   \centering
   \includegraphics[width=0.9\textwidth]{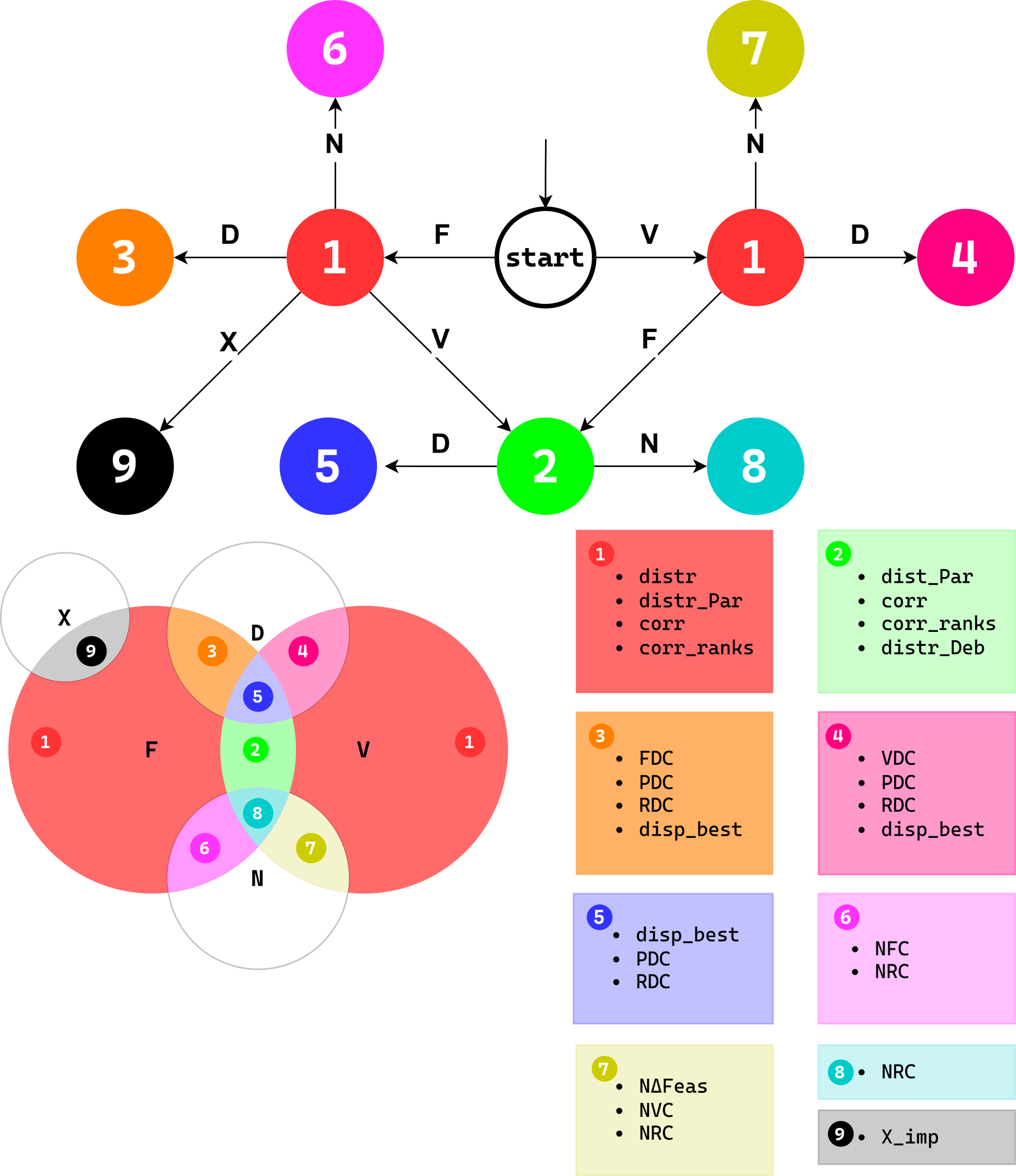}
   \caption{A diagrammatic summary of the landscape features currently available in \texttt{pyXla}. The transition diagram illustrates how richer analysis is provided as more input components are added. The Venn diagram depicts the various input component combinations, where the intersection between two circles indicates that both input components are present. Both the states of the transition diagram and sections of the Venn diagram are labelled with numbers. Each number label corresponds to a set of landscape features applicable to a given input component combination, as shown in the key, which is located to the right of the Venn diagram.}
   \label{fig:pyxla}
\end{figure}

The LA techniques implemented in the \texttt{pyXla} package are used to extract landscape features.
The features are categorised into four classes: statistical features, rank-based features, distance-based features, and neighbourhood-based features, discussed in \Cref{subsubsec:stats,subsubsec:ranking,subsubsec:dist,subsubsec:N}, respectively.
Figure \ref{fig:pyxla} provides a diagrammatic summary that depicts the inputs or combinations of inputs required by each feature.

\subsubsection{Statistical Features}\label{subsubsec:stats}
The features included in the first category are based on statistical techniques, and they are objective distribution (\texttt{distr\_F}), violation distribution (\texttt{distr\_V}), correlation of values (\texttt{corr}), correlation of ranks (\texttt{corr\_ranks}), and variable importance (\texttt{X\_imp}). The statistical features are listed in Table~\ref{tab:stats-feat}, giving a brief description of each feature, the inputs it requires, and the expected output. The numerical output of \texttt{distr\_F} and \texttt{distr\_V} consists of the following descriptive statistics: mean, median, minimum value, maximum value, first quartile, third quartile, standard deviation, skewness, and kurtosis.
We use these same statistics to characterise any distribution in \texttt{pyXla}.

\begin{table}[!htb]
\caption{Statistical landscape features included in \texttt{pyXla}. The symbols $\vee$ and $\wedge$ refer to conjunction and disjunction, respectively, of the presence of their operands, i.e., input types.}
\label{tab:stats-feat}
\begin{tabular}{l|l|p{8.6cm}}
\toprule

\textbf{Feature} & \textbf{Requires} & \textbf{Description}\\
\midrule

\texttt{distr\_F} 
& \F 
&\textbf{Name:} Objective distribution\\
&& \textbf{Visual output:} Histograms of function values and function ranks for each objective\\
&& \textbf{Numerical output:} Descriptive statistics\\
\midrule

\texttt{distr\_V} 
& \V 
& \textbf{Name:} Violation distribution \\
&& \textbf{Visual output:} Histogram of violation values and violation ranks for each constraint \\
&& \textbf{Numerical output:} Feasibility rate overall and per constraint, descriptive statistics\\
\midrule

\texttt{corr} 
& \F$~\vee~$\V 
& \textbf{Name:} Correlation of values\\
&& \textbf{Visual output:} Scatter plots of all objectives and violations against each other, with feasibility indicated per solution\\
&& \textbf{Numerical output:} Spearman's correlation coefficients for all pairs of sets of values split by feasibility\\
\midrule

\texttt{corr\_ranks} 
& \F$~\vee~$ \V 
& \textbf{Name:} Correlation of ranks\\
&& \textbf{Visual output:} Scatter plots of all objective, violation, Pareto, and Deb's ranks against each other, with feasibility indicated per solution\\
&& \textbf{Numerical output:} Spearman's correlation coefficients for all pairs of sets of ranks split by feasibility\\
\midrule

\texttt{X\_imp} 
& \F$~\wedge~$\X & \textbf{Name:} Variable importance \\
&& \textbf{Visual output:} For each objective, a bar chart of correlation values between each decision variable and the objective and a bar chart showing variable importance scores based on permutation feature importance using a linear model (Ridge or Lasso) fitted on decision variables and the objective, quantified by the coefficient of determination, $R^2$\\
&& \textbf{Numerical output:} Spearman's correlation coefficients of decision variables against each other and against the objectives, variable importance scores, and ranks\\
\bottomrule
\end{tabular}
\end{table}

\subsubsection{Rank-based Features}\label{subsubsec:ranking}
There are two rank-based features implemented.
First, the Pareto rank distribution (\texttt{distr\_Par}) corresponds to the ranking of solutions based on non-dominated sorting~\citep{goldberg1989}, as used, e.g., in NSGA-II~\citep{deb2002}. We apply it to objectives, violations, as well as their combination.
For data with a single objective, the Pareto rank simply corresponds to the ordering of objective values. In all cases, a lower rank is better.
Second, the \citet{deb2000efficient}'s feasibility rule ranking distribution (\texttt{distr\_Deb}) rank solutions by constraint violation(s) first, then objective value(s), ensuring that any feasible solution is always preferred over infeasible ones. These two features are categorised as rank-based since they involve ranking rules that are not merely statistical. They are listed in Table~\ref{tab:ranking-feat}, which gives a brief description of each feature, the inputs it requires, and the expected output.
The numerical output of \texttt{distr\_Par} and \texttt{distr\_Deb} consists of the same descriptive statistics as \texttt{distr\_F} and \texttt{distr\_V}.

\begin{table}[!htb]
\caption{Rank-based landscape features included in \texttt{pyXla}. The symbols $\vee$ and $\wedge$ refer to conjunction and disjunction, respectively, of the presence of their operands, i.e., input types.}
\label{tab:ranking-feat}
\begin{tabular}{l|l|p{8.9cm}}
\toprule

\textbf{Feature} & \textbf{Requires} & \textbf{Description}\\
\midrule

\texttt{distr\_Par} 
& \F$~\vee~$\V
& \textbf{Name:} Pareto rank distribution \\
&& \textbf{Visual output:} Histogram of Pareto ranks~\citep{goldberg1989} for objective values, violations, and the combination (treating violations as objectives) \\
&& \textbf{Numerical output:} Descriptive statistics\\
\midrule

\texttt{distr\_Deb} 
& \F$~\wedge~$\V
& \textbf{Name:} Deb's feasibility rule~\citep{deb2000efficient} ranks distribution\\
&& \textbf{Visual output:} Histogram of feasibility rules' ranking \\
&& \textbf{Numerical output:} Descriptive statistics\\
\bottomrule

\end{tabular}
\end{table}

\subsubsection{Distance-based Features}\label{subsubsec:dist}
Distance-based landscape features use distance information to characterise problems. They are:
fitness distance correlation (\texttt{FDC}),
violation distance correlation (\texttt{VDC}),
rank distance correlation (\texttt{RDC}),
pairwise distance correlation (\texttt{PDC}), and
dispersion of the best solutions (\texttt{disp\_best}). The distance-based features are listed in Table~\ref{tab:dist-feat}, which gives a brief description of each feature, the inputs it requires, and the expected output.

\begin{table}[!htb]
\caption{Distance-based landscape features included in \texttt{pyXla}. The symbols $\vee$ and $\wedge$ refer to conjunction and disjunction, respectively, of the presence of their operands, i.e., input types.}
\label{tab:dist-feat}
\begin{tabular}{l|l|p{8.6cm}}

\toprule
\textbf{Feature} & \textbf{Requires} & \textbf{Description}\\
\midrule

\texttt{FDC} 
& \F$~\wedge~$\D 
& \textbf{Name:} objective-distance correlation \\
&& \textbf{Visual output:} Scatter plots of objective values against distance to the nearest best solution in sample per objective, for all solutions or for feasible solutions only\\
&& \textbf{Numerical output:} Spearman's correlation coefficients per objective\\
\midrule

\texttt{VDC} 
& \V$~\wedge~$\D 
& \textbf{Name:} violation-distance correlation \\
&& \textbf{Visual output:} Scatter plots of violation values against distance to the nearest feasible solution in sample per constraint, for infeasible solutions only\\
&& \textbf{Numerical output:} Spearman's correlation coefficients per violation\\
\midrule

\texttt{RDC} 
& (\F$~\vee~$\V)$~\wedge~$\D 
& \textbf{Name:} rank-distance correlation \\
&& \textbf{Visual output:} Scatter plots of ranks against distance to the nearest best solution (with rank $1$) in the sample. Ranks based on Pareto ranks for objectives, violations, or the combination, or Deb feasibility rank.\\
&& \textbf{Numerical output:} Spearman's correlation coefficients per rank\\
\midrule
\texttt{PDC} 
& (\F$~\vee~$\V)$~\wedge~$\D 
& \textbf{Name:} pairwise distances correlation \\
&& \textbf{Visual output:} Scatter plots of pairwise distance in the variable space against distance in the objective space, violation space, and for each objective, constraint, and rank\\
&& \textbf{Numerical output:} Spearman's correlation coefficients for each scatter plot\\
\midrule

\texttt{disp\_best} 
& (\F$~\vee~$\V)$~\wedge~$\D & \textbf{Name:} Dispersion of best solutions \\
&& \textbf{Visual output:} Scatter plots showing the distribution of pairwise distances between solutions for increasing sample sizes of best solutions, where `best' is with respect to different objectives, constraints, and ranks.\\
&& \textbf{Numerical output:} dispersion metrics calculated using the smallest and largest sub-samples\\
\bottomrule

\end{tabular}
\end{table}

\subsubsection{Neighbourhood-based Features}\label{subsubsec:N}
The features in this category make use of neighbourhood information, and they include neighbouring solutions' objective (fitness) values correlation (\texttt{NFC}),
neighbouring solutions' violation values correlation (\texttt{NVC}),
neighbouring solutions' ranks correlation (\texttt{NRC}), and neighbouring change in feasibility (\texttt{N$\Delta$Feas}). The neighbourhood-based features are listed in Table~\ref{tab:neighbourhood-feat}, which gives a brief description of each feature, the inputs it requires, and the expected output.

\begin{table}[!htb]
\caption{Neighbourhood-based landscape features included in \texttt{pyXla}. The symbols $\vee$ and $\wedge$ refer to conjunction and disjunction, respectively, of the presence of their operands, i.e., input types.}
\label{tab:neighbourhood-feat}
\begin{tabular}{l|l|p{9.1cm}}

\toprule
\textbf{Feature} & \textbf{Requires} & \textbf{Description}\\
\midrule

\texttt{NFC} 
& \F$~\wedge~$\N 
& \textbf{Name:} Neighbouring solutions' objective values correlation\\ 
&& \textbf{Visual output:} A scatter plot of objective values between neighbours with a regression line for all solutions or feasible solutions only. The plot is divided by a broken line through the origin such that, assuming minimisation, points above the line are deteriorating neighbours, those below the line are improving neighbours, and those on the line are neutral neighbours.\\
&& \textbf{Numerical output:} Spearman's correlation coefficients\\
\midrule

\texttt{NVC} 
& \V$~\wedge~$\N 
& \textbf{Name:} Neighbouring solutions' violation values correlation\\
&& \textbf{Visual output:} A scatter plot of violation values between neighbours for each constraint, for infeasible solutions only, with a regression line. The plot is divided by a broken line through the origin such that points above the line are deteriorating neighbours, those below are improving neighbours, and those on the line are neutral neighbours.\\
&& \textbf{Numerical output:} Spearman's correlation coefficients\\
\midrule

\texttt{NRC} 
& (\F$~\vee~$\V)$~\wedge~$\N 
& \textbf{Name:} Neighbouring solutions' ranks correlation\\
&& \textbf{Visual output:} A scatter plot of ranks between neighbours. Ranks include objective ranks, violation ranks, Pareto ranks (for objectives, violations, or a combination of both), or Deb's feasibility rank. A regression line is fitted to the scatter plot. The plot is divided by a broken line through the origin such that points above the line are deteriorating neighbours, those below are improving neighbours, and those on the line are neutral neighbours.\\
&& \textbf{Numerical output:} Spearman's correlation coefficients\\
\midrule

\texttt{N$\Delta$Feas} 
& \V$~\wedge~$\N 
& \textbf{Name:} Neighbouring change in feasibility\\
&& \textbf{Visual output:} Bar charts of proportion of transitions from a feasible solution to an infeasible neighbour or from an infeasible solution to a feasible neighbour and proportion of transitions that do not result in a neighbour of dissimilar feasibility, for each constraint and their combination\\
&& \textbf{Numerical output}: Proportions indicating the extent to which feasibility regions are contiguous \\
\bottomrule


\end{tabular}
\end{table}

\subsection{The \texttt{pyXla} Package}
The four feature categories are implemented in a user-friendly open-source toolbox called \texttt{pyXla}. To the best of our knowledge, this is the only software that conceptually separates samples from features while supporting continuous and combinatorial domains, single and multiple objectives, as well as unconstrained and constrained problems within a unified framework. We deliberately limit the number of features to enhance explainability, as the package is not intended for automation of downstream tasks, but rather for problem understanding.
\texttt{pyXla} is published on the Python package index at: \href{https://pypi.org/project/pyxla}{https://pypi.org/project/pyxla}.
The main dependencies for the \texttt{pyXla} package are: \texttt{scipy}~\citep{2020SciPy-NMeth}\footnote{\href{https://scipy.org}{https://scipy.org}}, 
\texttt{pandas}~\citep{reback2020pandas,mckinney-proc-scipy-2010}\footnote{\href{https://pandas.pydata.org}{https://pandas.pydata.org}}, 
\texttt{moocore}\footnote{\href{https://github.com/multi-objective/moocore}{https://github.com/multi-objective/moocore}}, 
\texttt{hilbertcurve}\footnote{\href{https://github.com/galtay/hilbertcurve}{https://github.com/galtay/hilbertcurve}},
\texttt{matplotlib}~\citep{Hunter:2007}\footnote{\href{https://matplotlib.org}{https://matplotlib.org}}, 
\texttt{seaborn}~\citep{Waskom2021}\footnote{\href{https://seaborn.pydata.org}{https://seaborn.pydata.org}}, and 
\texttt{scikit-learn}~\citep{scikit-learn}\footnote{\href{https://scikit-learn.org/}{https://scikit-learn.org/}}.
The source code and all reproducible experiments are also made available under the MIT license on GitLab at: \href{https://gitlab.com/aliefooghe/pyxla}{https://gitlab.com/aliefooghe/pyxla}. Accompanying documentation is also published to facilitate adoption: \href{https://pyxla.readthedocs.io}{https://pyxla.readthedocs.io}.

\section{Illustrative Use Cases of Explainable Landscape Analysis}
\label{sec:examples}
This section illustrates the flexibility, generality, and explainability of the proposed framework using selected problem instances. \Cref{subsec:problems} defines the selected problem instances while \Cref{subsec:feat-illustrate} discusses the outputs of the features in each category for the different use cases.

\subsection{Selected Problem Instances}\label{subsec:problems}
To illustrate the flexibility, generality, and explainability of \texttt{pyXla} for problem understanding, a set of problem instances is curated to contain single-objective and multi-objective, as well as constrained and unconstrained problems from both combinatorial and continuous domains. Additionally, the problems are of varying complexity, such that both easy and difficult problems are included. \Cref{tab:problems} summarises the selected problem instances, indicating the problem representation and the sampling strategy.
We discuss each of these problems below.

\begin{table}[!t]
\caption{Selected problem instances, with
$n$ denoting the problem size.
For NK landscape instances, 
$g$ are random lookup tables that draw real numbers uniformly from the range [0, 1], see \Cref{subsubsec:nk} for details.
For Hilbert curve sampling, the sample size is $100$.
}
\label{tab:problems}
\begin{tabular}{l|llll}
\toprule

\textbf{Problem} & \textbf{Domain} & \textbf{Objective(s)} & \textbf{Constraint(s)} &\textbf{Sampling}\\
\midrule
Sphere ($n\!=\!1$)
& Cont.
& \cref{eq:sphere}
& eqs.~(\ref{eq:sphere-constraint}), (\ref{eq:sine-constraint})
& Hilbert curve\\ 
%
Deceptive ($n\!=\!1$)
& Cont.
& \cref{eq:deceptive}
& \cref{eq:deceptive-v0}
& Hilbert curve\\ 
%
Easom ($n\!=\!2$)
& Cont.
& \cref{eq:easom}
& None
& Hilbert curve\\ 
%
Easy knapsack ($n\!=\!10$)
& Combin.
& Profit
& Weight
& Exhaustive\\
%
Hard knapsack ($n\!=\!10$)
& Combin.
& Profit
& Weight
& Exhaustive\\
%
NK ($n\!=\!14, k\!=\!1$)
& Combin.
& $g$
& None
& Exhaustive\\
%
NK ($n\!=\!14, k\!=\!2$)
& Combin.
& $g$
& None
& Exhaustive\\
%
NK ($n\!=\!14, k\!=\!4$) 
& Combin.
& $g$
& None
& Exhaustive\\
%
NK ($n\!=\!14, k\!=\!2$) 
& Combin.
& $g_0, g_1, g_2$
& None
& Exhaustive\\
XOR gate NN ($n\!=\!9$) 
& Cont.
& \cref{eq:bce-loss}
& None
& SGD trace\\
XOR gate NN ($n\!=\!9$) 
& Cont.
& \cref{eq:bce-loss}
& None
& Random\\
\bottomrule
\end{tabular}
\end{table}

\subsubsection{Constrained Sphere Problem}
\begin{figure}[ht!]
   \begin{subfigure}{0.30\textwidth}
      \includegraphics{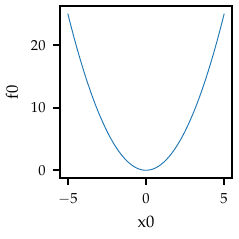}
      \caption{Objective (\ref{eq:sphere})}
      \label{fig:sphere-2d}
   \end{subfigure}
   \begin{subfigure}{0.30\textwidth}
      \includegraphics{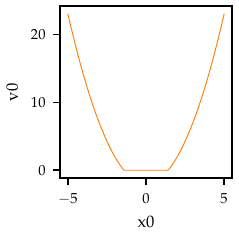}
      \caption{Constraint~(\ref{eq:sphere-constraint})}
      \label{fig:sphere-v0}
   \end{subfigure}
   \begin{subfigure}{0.30\textwidth}
      \includegraphics{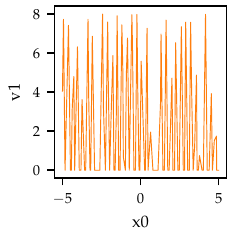}
      \caption{Constraint~(\ref{eq:sine-constraint})}
      \label{fig:sphere-v1}
   \end{subfigure}
   \caption{Line plots of the constrained sphere problem objective and constraints violation values.}
\end{figure}

The constrained sphere problem uses the sphere function to generate objective values, 
defined as:
\begin{equation}\label{eq:sphere}
   f_1(x) = x^2.
\end{equation}
A sample of 100 $1$-dimensional points within a bounding box $[-5, 5]$ is obtained using Hilbert curve sampling to ensure adequate coverage of the search space.
The points are then evaluated on the sphere function to generate the objective values. 
The objective is to be minimised. 
The problem additionally includes two constraints, defined as:
\begin{equation}\label{eq:sphere-constraint}
   v_0(x) = x^2 - 2 \leq 0,~\text{and}
\end{equation}
\begin{equation}\label{eq:sine-constraint}
   v_1(x) = 8 \cdot \sin(20 \cdot x) \leq 0.
\end{equation}
The objective and violation values with respect to Constraints (\ref{eq:sphere-constraint}) and (\ref{eq:sine-constraint}) are visualised in \Cref{fig:sphere-2d,fig:sphere-v0,fig:sphere-v1}, respectively.

\subsubsection{Constrained Deceptive Problem}
\begin{figure}[ht!]
   \centering
   \begin{subfigure}{0.30\textwidth}
      \includegraphics[width=\textwidth]{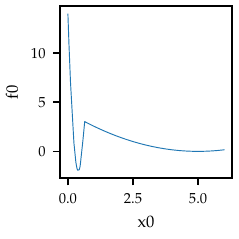}
      \caption{Objective (\ref{eq:deceptive})}
      \label{fig:deceptive-f0}
   \end{subfigure}
   \begin{subfigure}{0.30\textwidth}
      \includegraphics[width=\textwidth]{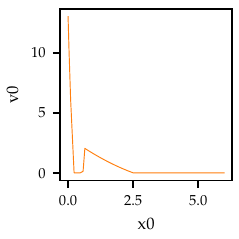}
      \caption{Constraint~(\ref{eq:deceptive-v0})}
      \label{fig:deceptive-v0}
   \end{subfigure}
   \caption{Line plots of the constrained deceptive problem's objective~($f0$) and constraint~($v0$).}
   \label{fig:deceptive}
\end{figure}

The constrained deceptive problem is defined as the minimisation of a piecewise function: 
\begin{equation}\label{eq:deceptive}
   f_2(x) = 
   \begin{cases}
        (0.4x - 2)^2 & \text{if } x > 0.625\\
        (10x - 4)^2 - 2 & \text{otherwise}.
    \end{cases}
\end{equation}
A sample of 100 1-dimensional solutions is sampled using the Hilbert curve within the bounding box $[-5, 5]$. The constraint is defined in terms of the objective function~(\ref{eq:deceptive}) as:
\begin{equation}\label{eq:deceptive-v0}
   v_0(x) = f_2(x) - 1 \leq 0.
\end{equation}
The objective and violation values are visualised in \Cref{fig:deceptive}. The problem is designed to be deceptive, as it has two local minima with the global minimum in a narrow basin.

\subsubsection{(Unconstrained) Easom Function}
\begin{figure}[h!]
   \centering
   \includegraphics{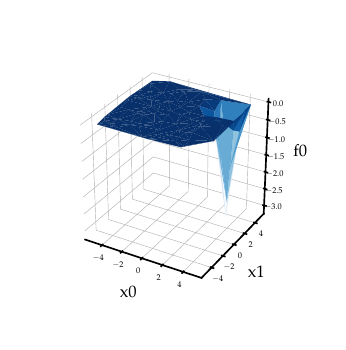}
   \caption{Surface plot of the Easom function.}
   \label{fig:easom}
\end{figure}

Similarly, we use the Easom function~\citep{easom1990survey} to generate objective values from a sample with 100 2-dimensional points sampled using the Hilbert curve within the bounding box $[-5, 5]$. The Easom function is defined as:
\begin{equation}\label{eq:easom}
   f_3(x_1, x_2) = -\cos(x_1) \cdot \cos(x_2) \cdot e^{-(x_1 - \pi)^2 - (x_2 - \pi)^2},
\end{equation}
and is visualised in \Cref{fig:easom}.
As shown in \Cref{fig:easom}, the Easom function is unimodal and has a high degree of neutrality~\citep{van2017characterising}. This example is unconstrained.

\subsubsection{Knapsack Instances}
The knapsack is a combinatorial optimisation problem where out of a set of $n$ items, each associated with a profit and weight, a combination of items has to be chosen such that the profit is maximised, and the resulting cumulative weight does not exceed the maximum weight allowable~\citep{pisinger2005hard}.
Knapsack problem instances are chosen to represent discrete optimisation problems and serve as an illustration for constrained problems.

Two knapsack problem instances are chosen: \emph{easy} and \emph{hard} instances. The \emph{easy} instance is created by specifying a large maximum weight constraint, such that its search space has a high feasibility rate, and where the profits and the weights are uncorrelated. Problems where profit and weight are uncorrelated are generally easier to solve~\citep{pisinger2005hard,han2010hard}.
%
By contrast, a \emph{hard} knapsack instance is created by specifying a small maximum weight constraint corresponding to a search space with a low feasibility rate. Additionally, the instance is configured such that there is a strong correlation between the profits and the weights. Instances where profit and weight are strongly correlated are generally harder to solve~\citep{pisinger2005hard,han2010hard}, as higher profit items are weightier. 

\subsubsection{NK Landscape Instances}\label{subsubsec:nk}
The NK model was introduced by ~\cite{kauffmanGeneralTheoryAdaptive1987} to capture key aspects of biological evolution. Since then, it has since been widely adopted  in combinatorial optimisation and in many other fields.
In particular, any pseudo-Boolean function can be represented as an NK landscape, making the model a general framework for binary optimisation. Several extensions have also been proposed, including multi-objective NK landscapes~\citep{aguirre2007,VerLieJou2013ejor}. 
Following~\cite{pelikanNKLandscapesProblem2010}, 
an NK landscape consists of the components below:
\begin{enumerate}
   \item The number of bits $n$, such that solutions in the search space are of the form $(x_0, x_1, ..., x_{n-1})$;
   \item The number of interactions per bit $k$;
   \item A set of $k$ interactions $\mathcal{I}(x_i)$ for every bit in $\{x_0, x_1, ..., x_{n-1}\}$, such that the contribution of each $x_i$ to the fitness of a given configuration depends on both $x_i$ and its $k$ interacting variables;
   \item A sub-function(s) $g_i$ defining a real value for each of the combinations of values of $x_i$ and $\mathcal{I}(x_i)$, with $g_i$ being a lookup table. The sub-function $g_i$ assigns an objective value by drawing a random real number uniformly from $[0, 1]$.
\end{enumerate}
Three unconstrained single-objective and one constrained multi-objective NK landscape instances are included in the testing suite. Each instance has a dimension of $n = 14$ with varying difficulty of optimisation defined by its $k$ values. The parameter $k$ influences the ruggedness of the landscape~\citep{weinberger1991,ochoaStudyNKLandscapes2008}. The easiest of the single-objective instances has $k = 1$, the next one has $k=2$, and the hardest has $k=4$. Increasing the value of $k$ makes the landscape more rugged and thus harder to solve. The multi-objective instance has $k=2$ with three objectives (hence three lookup tables $g_0, g_1, g_2$).

\subsubsection{Simulating XOR Gate using a Neural Network}

\begin{figure}[t!]
   \centering
   \includegraphics[width=0.8\textwidth]{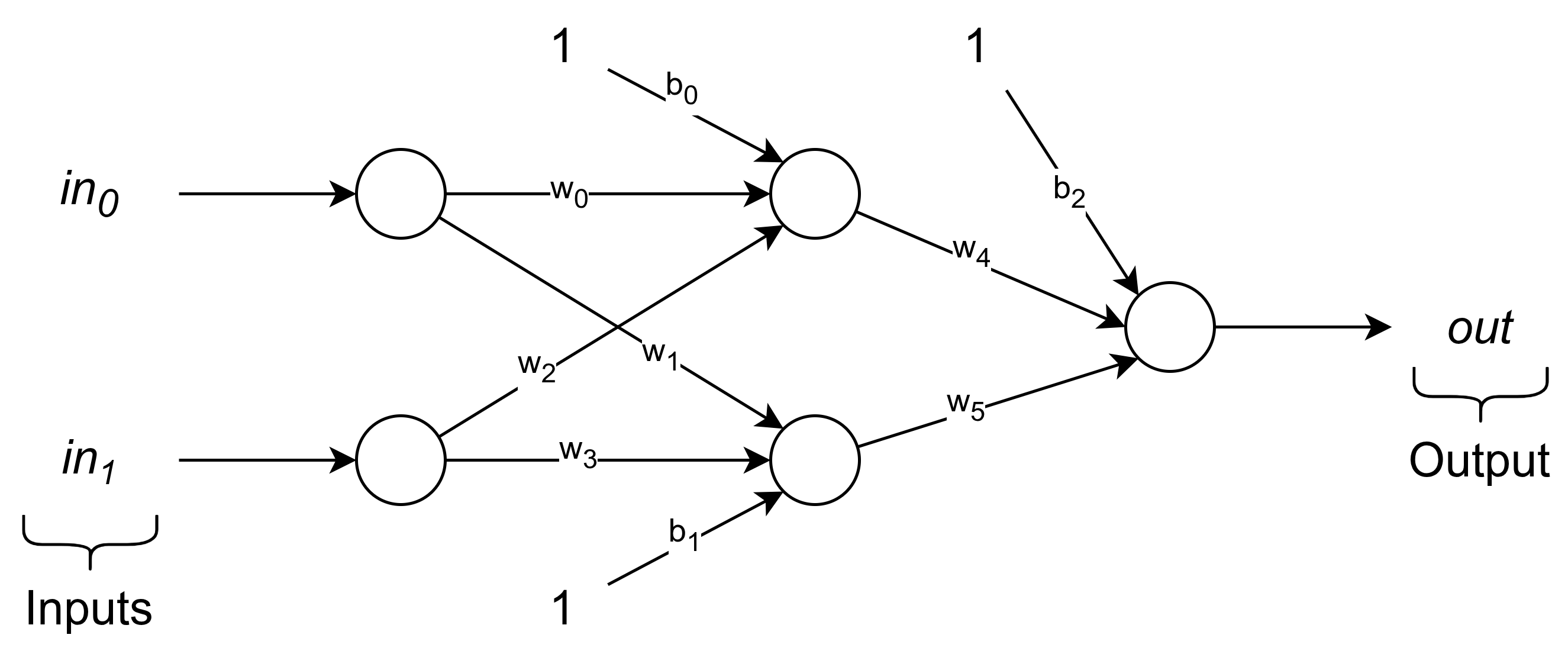}
   \caption{Neural network simulating the XOR gate. The weights and biases are labelled as $w_i$ and $b_i$, respectively. The variables $in_1$ and $in_2$ correspond to columns Input A and Input B in \Cref{tab:xor}. The biases are multiplied by a constant input of 1.}
   \label{fig:nn}
\end{figure}

A neural network with 9 parameters (weights and biases) is optimised to simulate an XOR gate.
A neural network is a neuroscience-inspired function approximator consisting of a composition of functions with trainable parameters~\citep{Goodfellow-et-al-2016}.
The neural network depicted in \Cref{fig:nn} is used.
In the hidden layer, $\tanh(x)$ is used as the activation function,
\begin{equation}
    \tanh(x) = \frac{e^x - e^{-x}}{e^x + e^{-x}}.
\end{equation}
The sigmoid function is used to convert logits at the output neuron to probabilities, which are converted to Boolean logic output using 0.5 as the threshold;
\begin{equation}
    sigmoid(x) = \frac{1}{1 + e^{-x}}.
\end{equation}
For this problem, two scenarios are comparatively analysed: (1) the trajectory of stochastic gradient descent (SGD) on the parameter space when optimising the neural network over 1000 epochs to fit the truth table (\Cref{tab:xor}) of the XOR gate; (2) randomly sampling 1000 neural network parameter configurations.
In both scenarios, the objective function is the binary cross-entropy loss function show in \Cref{eq:bce-loss}:
\begin{equation}\label{eq:bce-loss}
    loss = \frac{1}{N}\sum^N_{n=1}[y_n\log(p_n) + (1-y_n)\log(1-p_n)],
\end{equation}
where $y_n$ is the expected truth value for a given set of XOR operands, $p_n$ is the truth value predicted at the output layer, and summation is over the four entries of the XOR truth table.
Before optimisation in scenario (1), the weights of the neural network are initialised by sampling from the Xavier normal distribution~\citep{glorot2010understanding} $\mathcal{N}(0, \sigma^2)$ where,
\begin{equation}\label{eq:xavier-var}
    \sigma^2 = \frac{2}{fan_{in} + fan_{out}}.
\end{equation}
In \Cref{eq:xavier-var}, $fan_{in}$ refers to the number of incoming connections into a given neuron and $fan_{out}$ the number of outgoing connections.
The biases are initialised by randomly drawing from the uniform distribution in the range [0, 1].
In scenario (2), weights and biases are sampled from the same distributions as those used in scenario (1), respectively, for fairness in comparison.
The same seed is used for the random number generators in both scenarios.
We thus study the landscape of optimising the six weights and three biases of the neural network under both scenarios.

\begin{table}[!t]
\centering
\caption{XOR gate truth table.}
\label{tab:xor}
\begin{tabular}{ll|l}
\toprule
\textbf{Input $A$} & \textbf{Input $B$} & \textbf{Output ($A\oplus B$)}\\
\midrule
False
& False
& False\\
False
& True
& True\\
True
& False
& True\\
True
& True
& False\\
\bottomrule
\end{tabular}
\end{table}

\subsection{\texttt{pyXla} in Action}
\label{subsec:feat-illustrate}
This section illustrates how each feature in the proposed XLA framework produces explainable output using the selected problem instances presented in \Cref{subsec:problems}. \Cref{subsubsec:stats-feat,subsubsec:rank-feat,subsubsec:dist-feat,subsubsec:neighbourhood-feat} present the illustrative use cases of the features, organised by category.
For reproducibility, the examples below are also provided as a Jupyter notebook in supplementary material at: \href{https://gitlab.com/aliefooghe/pyxla/-/tree/main/docs/source/notebooks}{https://gitlab.com/aliefooghe/pyxla/-/tree/main/docs/source/notebooks}. All figures and numbers below were produced using \texttt{pyXla 1.0.0}.

\subsubsection{Statistical Features}\label{subsubsec:stats-feat}
We start with illustrative use cases for the statistical features.

\begin{figure}[htb!]
\centering
   \begin{subfigure}{0.3\textwidth}
      \includegraphics[width=\textwidth]{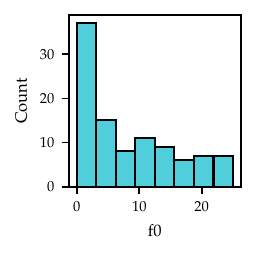}
      \caption{Sphere function}
      \label{fig:distr-f-sphere}
   \end{subfigure}
   \begin{subfigure}{0.3\textwidth}
      \includegraphics[width=\textwidth]{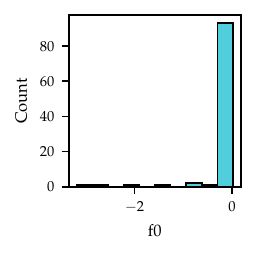}
      \caption{Easom function}
      \label{fig:distr-f-easom}
   \end{subfigure}
   \caption{Output of \texttt{distr\_F}: histogram of objective values for the constrained sphere problem and the Easom function.}
\end{figure}

\paragraph{Objective Distribution (\texttt{distr\_F}):}
The \texttt{distr\_F} feature captures the spread of objective values and computes some descriptive statistics of the objective values.
Comparing the sphere function and the Easom function, it is expected that the spread of the objective values will be different, as the sphere function is parabolic, whilst the Easom function has some degree of neutrality. The sphere and Easom functions are visualised in Figures \ref{fig:sphere-2d} and \ref{fig:easom}, respectively. The visual output of the \texttt{distr\_F} on the constrained sphere problem and the Easom function problem is depicted in Figures \ref{fig:distr-f-sphere} and \ref{fig:distr-f-easom}. The spread of the objective values is distinctive for each problem. \Cref{fig:distr-f-sphere} shows that the distribution of the sphere function objective values is that of the right tail of the normal distribution. \Cref{fig:distr-f-easom} shows that the Easom function has neutrality, since many solutions have nearly the same objective values.
The numerical outputs, as presented in \Cref{tab:distr-f}, show that the objective values of the two functions have different statistical properties. For instance, the sphere function has a skewness of 0.6142 and platykurtic kurtosis. On the other hand, the distribution of objective values of the Easom function has a skewness of -4.6114 and is leptokurtic (19.8784), indicating the presence of outliers, as is evident from \Cref{fig:easom}.

\begin{figure}[htb!]
\centering
   \begin{subfigure}{0.3\textwidth}
      \includegraphics{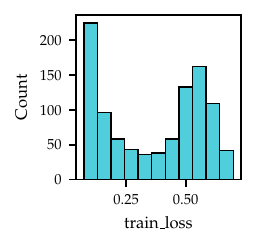}
      \caption{SGD trajectory}
      \label{fig:distr-f-xor-nn-trained}
   \end{subfigure}
   \begin{subfigure}{0.3\textwidth}
      \includegraphics{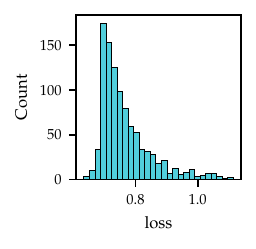}
      \caption{Random sampling}
      \label{fig:distr-f-xor-nn-sampled}
   \end{subfigure}
   \caption{Output of \texttt{distr\_F}: histogram of objective values for simulating the XOR gate using a neural network: (\subref{fig:distr-f-xor-nn-trained}) corresponds to the distribution of loss values for an SGD trajectory in parameter space, (\subref{fig:distr-f-xor-nn-sampled}) shows the distribution of loss values when the neural network parameters are randomly sampled.}
\end{figure}

\Cref{fig:distr-f-xor-nn-trained,fig:distr-f-xor-nn-sampled} show the output of \texttt{distr\_F} for the two scenarios simulating the  XOR gate using a neural network.
Most of the loss values in scenario (1), where the neural network is trained, are better (lower) than the best loss value in scenario (2), where parameters are randomly sampled.
This indicates that SGD effectively optimises the problem, considering scenario (2) as a baseline.

\begin{table}[htb!]
   \centering
   \caption{Numerical output of \texttt{distr\_F} for the constrained sphere problem and the Easom function.}
   \label{tab:distr-f}
   \begin{tabular}{l|rr}
      \toprule
      \textbf{Statistic} & \textbf{Sphere} & \textbf{Easom}\\
      \midrule
      Minimum & 0.0015 & -1.5046\\
      Maximum & 25.0 & 0.0277\\
      Mean & 8.7426 & -0.0606\\
      Median & 6.5597 & 6.5705e-33\\
      Q1 & 1.8607 & -1.4481e-16\\
      Q3 & 14.873 & 1.0137e-08\\
      Standard deviation & 7.6791 & 0.269\\
      Skewness & 0.6142 & -4.6114\\
      Kurtosis & -0.9287 & 19.8784\\
      \bottomrule
   \end{tabular}
\end{table}

\begin{figure}[t!]
\centering
   \begin{subfigure}{0.3\textwidth}
      \includegraphics{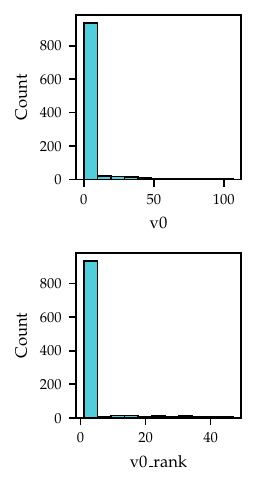}
      \caption{Easy knapsack instance}
      \label{fig:distr-v-easy-kp}
   \end{subfigure}
   \begin{subfigure}{0.3\textwidth}
      \includegraphics{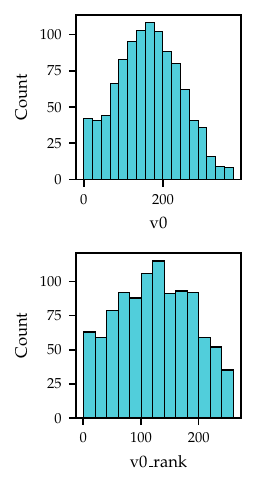}
      \caption{Hard knapsack instance}
      \label{fig:distr-v-hard-kp}
   \end{subfigure}
   \caption{Output of \texttt{distr\_V}: histogram of violation values (top row) and ranks (bottom row) for the easy and hard knapsack instances.}
\end{figure}

\paragraph{Violation Distribution (\texttt{distr\_V}):} The \texttt{distr\_V} feature is similar to the \texttt{distr\_F} except that it is applied to the violation values instead of the objective values.
The \texttt{distr\_V} feature is used here to compare the easy knapsack instance to the hard one. The numerical output \texttt{distr\_V} indicates that the feasibility rates of the easy and hard knapsack instances are 88.18\% and 0.0977\%, respectively. \Cref{fig:distr-v-easy-kp,fig:distr-v-hard-kp} show the distribution of violation values (top row) and ranks (bottom row) corresponding to the feasibility rates. The majority of solutions of the easy knapsack instance have a violation value of zero since most solutions are feasible; hence, the spread of both values and ranks is skewed to the left. The majority of solutions of the hard knapsack instance have violation values greater than zero since most solutions are infeasible. As a result, the violation values and ranks are normally distributed.

\begin{figure}[ht!]
   \centering
   \includegraphics{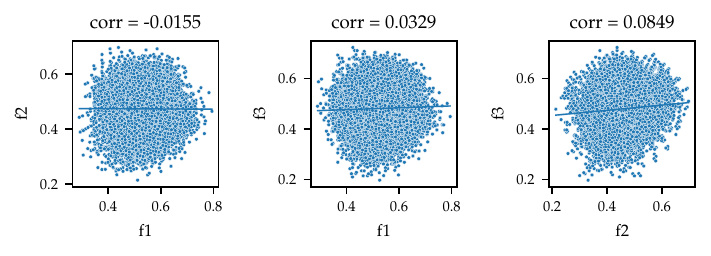}
   \caption{Output of \texttt{corr}: scatter plots of pairs of objectives for an NK instance with 3 objectives}
   \label{fig:corr}
\end{figure}

\paragraph{Correlation of Values (\texttt{corr}):} 
The \texttt{corr} feature~\citep{liefooghe_tevc2020} is used to analyse a multi-objective NK instance with $n=14$, $k=2$, and three objectives, resulting in the output in \Cref{fig:corr}. The correlation plots and the Spearman's correlation coefficient values in \Cref{fig:corr} show that the objective pairs are weakly correlated. This is expected since the objective values are randomly generated. The \texttt{corr} feature is thus useful for analysis of multi-objective problems.

\paragraph{Correlation of Ranks (\texttt{corr\_ranks}):}
The \texttt{corr\_ranks} feature is used to analyse the constrained sphere problem. \Cref{fig:corr-ranks} shows the visual output annotated with the numerical outputs. The ranks for Constraint v0~(\ref{eq:sphere-constraint}), are highly correlated with every other ranking, except those for Constraint v1~(\ref{eq:sine-constraint}). Constraint v1~(\ref{eq:sine-constraint}) being sinusoidal, it is weakly correlated with all the other ranks. Weak correlation with respect to the objective influences problem difficulty, as the solutions with the best objective values may be infeasible since there is no relationship between objective and violation values. By contrast, a strong positive correlation with the objective, as is the case for Constraint v0~(\ref{eq:sphere-constraint}), means there is a high likelihood that solutions with the best (smallest) objective values have lower violation values, and thus a higher likelihood of feasibility.

\begin{figure}[ht!]
   \centering
   \includegraphics{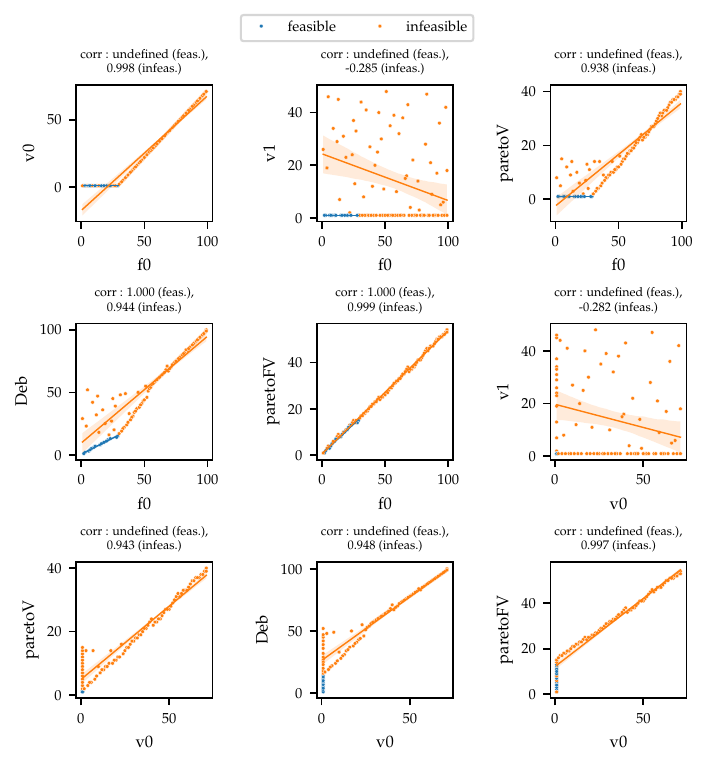}
   \caption{Output of \texttt{corr\_ranks}: scatter plots of pairs of ranks for the constrained sphere problem, split by feasibility; \emph{feas.} is an abbreviation for feasible.}
   \label{fig:corr-ranks}
\end{figure}
\begin{figure}[ht!]
   \ContinuedFloat
   \centering
   \includegraphics{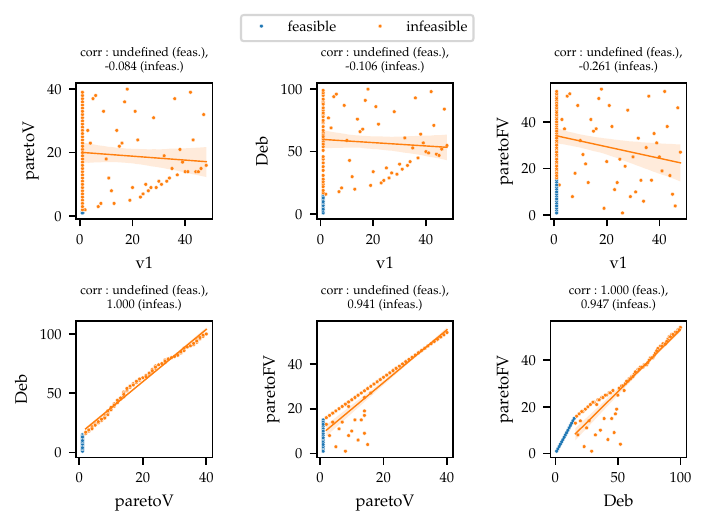}
   \caption{(Continued) Output of \texttt{corr\_ranks}: scatter plots of pairs of ranks for the constrained sphere problem, split by feasibility; \emph{feas.} is an abbreviation for feasible.}
   \label{fig:corr-ranks-continued}
\end{figure}

\paragraph{Variable importance (\texttt{X\_imp}):}
The \texttt{X\_imp} feature is applied to the scenarios simulating the XOR gate using a neural network and to the NK problem instances.

\begin{figure}[htb!]
\centering
   \begin{subfigure}{0.4\textwidth}
      \includegraphics[width=\textwidth]{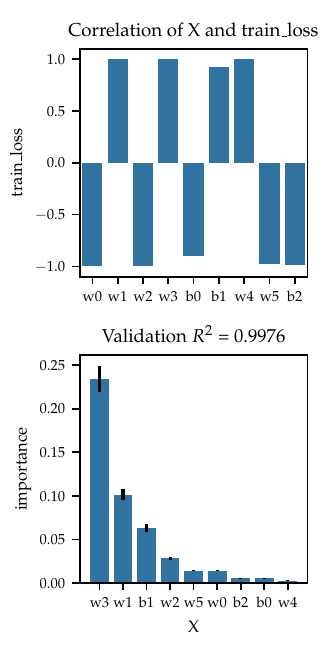}
      \caption{SGD trajectory}
      \label{fig:x-imp-xor-nn-trained}
   \end{subfigure}
   \begin{subfigure}{0.4\textwidth}
      \includegraphics[width=\textwidth]{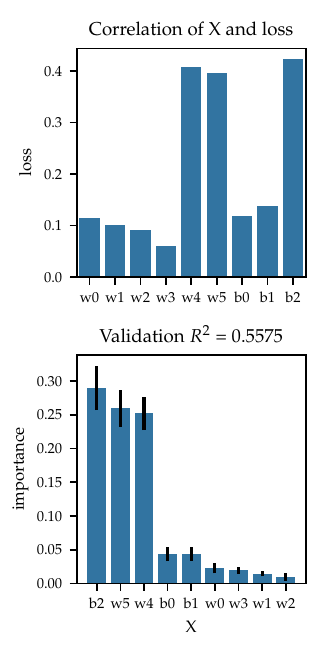}
      \caption{Random sampling}
      \label{fig:x-imp-xor-nn-sampled}
   \end{subfigure}
   \caption{Output of \texttt{X\_imp} for scenarios simulating the XOR gate using a neural network: (\subref{fig:x-imp-xor-nn-trained}) corresponds to the SGD trajectory in parameter space, and (\subref{fig:x-imp-xor-nn-sampled}) corresponds to the scenario where neural network parameters are randomly sampled.  The bar charts in the top row show the correlation values between each variable and the objective. The bar charts in the bottom row show the permutation feature importance scores.}
\end{figure}

\Cref{fig:x-imp-xor-nn-trained,fig:x-imp-xor-nn-sampled} show the output for the scenarios simulating the XOR gate using a neural network.
The variables whose importance is computed correspond to the weights and biases that can be mapped to the labels in \Cref{fig:nn}.
As expected, the relative importance of the weights and biases is substantially different in each scenario, since the first scenario involved the SGD trajectory, while the second originates from random sampling.
For scenario (1), the two most important weights are $w_3$ and $w_1$, which correspond to the input weights of the second hidden neuron (see Figure~\ref{fig:nn}). Combined with the bias $b_1$ (third most important parameter), these three parameters indicate that the second neuron is likely responsible for the positive activation when either of the inputs is 1. For scenario (2), on the other hand, the weights and bias associated with the output neuron are the most important ($b_2, w_5, w_4$). This makes sense: since no training is taking place, the hidden neurons act as random projections. The output neuron is directly responsible for producing the output, as such, with no mechanism to delegate function representation to the hidden layer, it automatically becomes the most correlated with the objective.

\begin{figure}[ht!]
   \begin{subfigure}{0.295\textwidth}
      \includegraphics[width=\textwidth]{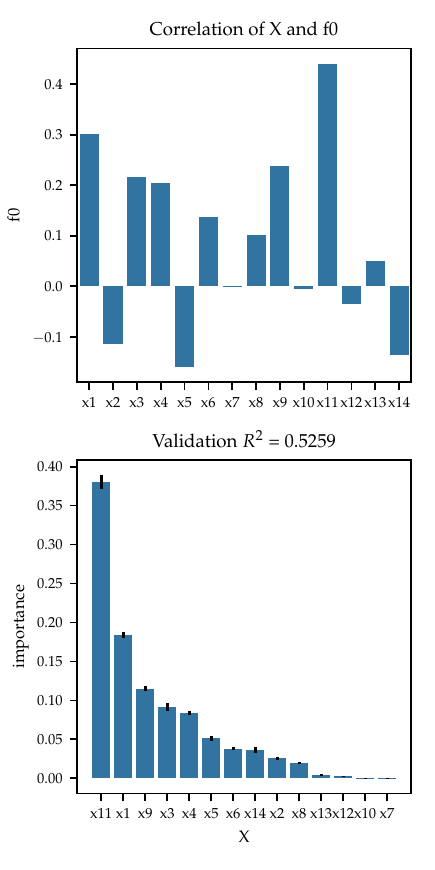}
      \caption{$k=1$}
      \label{fig:x-imp-nk-k1}  
   \end{subfigure}
   \hfill
   \begin{subfigure}{0.295\textwidth}
      \includegraphics[width=\textwidth]{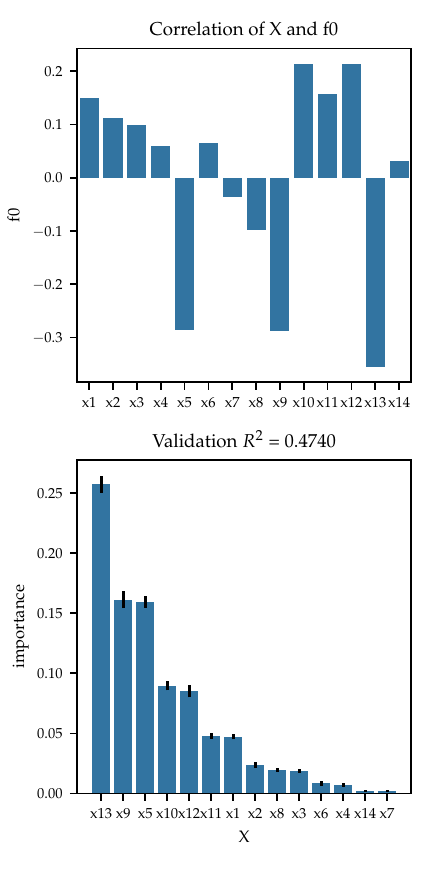}
      \caption{$k=2$}
      \label{fig:x-imp-nk-k2}  
   \end{subfigure}
   \hfill
   \begin{subfigure}{0.295\textwidth}
      \includegraphics[width=\textwidth]{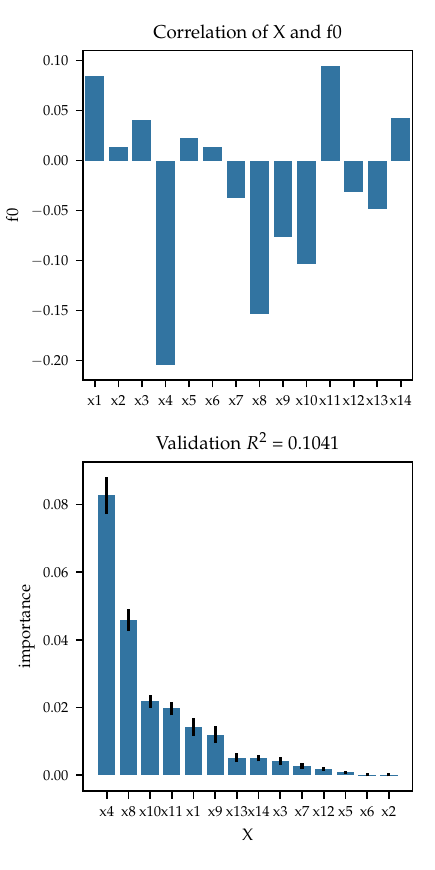}
      \caption{$k=4$}
      \label{fig:x-imp-nk-k4}  
   \end{subfigure}
   \caption{Visual output of the \texttt{X\_imp} feature applied to the NK problem instances. The bar charts in the top row show the correlation values between each variable and the objective. The bar charts in the bottom row show the permutation feature importance scores.}
\end{figure}

\Cref{fig:x-imp-nk-k1,fig:x-imp-nk-k2,fig:x-imp-nk-k4} show the visual output for NK landscapes with $k=1$, $k=2$, and $k=4$, respectively. Permutation feature importance is evaluated using a Ridge regression model with $\alpha=1$. It should be noted that there is agreement between Spearman's correlation coefficients and the permutation feature importance scores. Additionally, in general, the higher the value of $k$, the lower the correlation with the objective, and the lower the permutation feature importance score. This is consistent with the property of NK landscapes increasing in complexity with higher values of $k$. As complexity increases, the individual contribution of each variable reduces as its contribution depends not only on the variable itself but also on other variables, a dependency which worsens with increasing values of $k$.

\subsubsection{Rank-based Features}\label{subsubsec:rank-feat}
In this section, the illustrative use cases for the rank-based features are presented.

\begin{figure}[ht!]
    \centering
    \begin{subfigure}{\textwidth}
      \centering
      \includegraphics{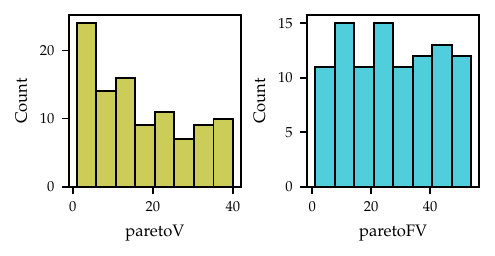}
      \caption{Constrained sphere problem}
      \label{fig:distr-Par-sphere}
    \end{subfigure}
    \\
    \begin{subfigure}{\textwidth}
      \centering
      \includegraphics{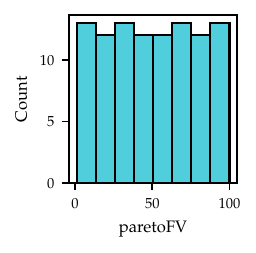}
      \caption{Constrained deceptive problem}
      \label{fig:distr-Par-deceptive}
    \end{subfigure}
    \caption{Histogram of Pareto ranks of objective and violation functions.}
\end{figure}

\paragraph{Pareto Rank Distribution (\texttt{distr\_Par}):}
This feature produces a histogram of Pareto ranks and descriptive statistics (similar to those of \texttt{distr\_F}) on the ranks.
\Cref{fig:distr-Par-sphere,fig:distr-Par-deceptive} show the visual output of the \texttt{distr\_Par} feature for the constrained sphere problem and the constrained deceptive problem. \Cref{tab:distr-Par-sphere,tab:distr-Par-deceptive} present the corresponding numerical outputs. The constrained sphere problem has Pareto ranks for constraints only, $paretoV$, and for both objectives and constraints, $paretoFV$. The rank $paretoV$ is only defined where there are multiple constraint functions. The constrained deceptive problem has only $paretoFV$, since it has only one violation and one objective. The Pareto ranks for the constrained deceptive problem are spread out evenly, while for the constrained sphere problem, some solutions share the same ranking. Thus, the maximum rank for the constrained deceptive problem is 100, while for the constrained sphere problem, the maximum rank is 52. The difference in the range of ranks for the two problems can be explained by the constrained sphere problem having two constraints (\Cref{eq:sphere-constraint,eq:sine-constraint}), hence larger Pareto fronts when considering violations as objectives ($paretoFV$). The constrained deceptive problem has one constraint (\Cref{eq:deceptive-v0}), which is correlated with the objective. Therefore, the ranking of violation values and objective values is identical, resulting in unit-size Pareto fronts and uniformly distributed $paretoFV$ ranks.

\begin{table}[ht]
   \centering
   \caption{Output of \texttt{distr\_Par}: descriptive statistics of the Pareto rank of objective and violation values for the constrained sphere problem.}
   \label{tab:distr-Par-sphere}
   \begin{subtable}{0.45\textwidth}
   \centering
       \begin{tabular}{l|r}
          \toprule
          \textbf{Statistic} & \textbf{Value}\\
          \midrule
          $paretoFV$ minimum & 1\\
          $paretoFV$ maximum & 52\\
          $paretoFV$ mean & 25.99\\
          $paretoFV$ median & 25.0\\
          $paretoFV$ Q1 & 12.0\\
          $paretoFV$ Q3 & 39.75\\
          $paretoFV$ sd & 15.3264\\
          $paretoFV$ skewness & 0.04444\\
          $paretoFV$ kurtosis & -1.2840 \\
          \bottomrule
       \end{tabular}
   \end{subtable}
   \begin{subtable}{0.45\textwidth}
       \centering
       \begin{tabular}{l|r}
          \toprule
          \textbf{Statistic} & \textbf{Value}\\
          \midrule        
          $paretoV$ minimum & 1\\
          $paretoV$ maximum & 38\\
          $paretoV$ mean & 16.07\\
          $paretoV$ median & 14.0\\
          $paretoV$ Q1 & 6.25\\
          $paretoV$ Q3 & 25.75\\
          $paretoV$ sd & 11.4629\\
          $paretoV$ skewness & 0.2888\\
          $paretoV$ kurtosis & -1.1496\\
          \bottomrule
        \end{tabular}
   \end{subtable}
\end{table}

\begin{table}[ht!]
   \centering
   \caption{Output of \texttt{distr\_Par}: descriptive statistics of the Pareto rank of objective and violation values for the constrained deceptive problem.}
   \label{tab:distr-Par-deceptive}
   \begin{tabular}{l|r}
      \toprule
      \textbf{Statistic} & \textbf{Value}\\
      \midrule
      $paretoFV$ minimum & 1\\
      $paretoFV$ maximum & 100\\
      $paretoFV$ mean & 50.5\\
      $paretoFV$ median & 50.5\\
      $paretoFV$ Q1 & 25.25\\
      $paretoFV$ Q3 & 75.75\\
      $paretoFV$ sd & 29.0115\\
      $paretoFV$ skewness & 0.0\\
      $paretoFV$ kurtosis & -1.2002\\
      \bottomrule
   \end{tabular}
\end{table}

\begin{figure}[ht!]
   \centering
   \begin{subfigure}{0.3\textwidth}
      \centering
      \includegraphics[width=\textwidth]{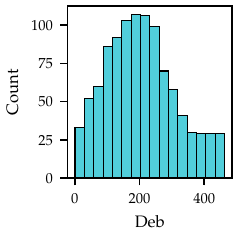}
      \caption{Easy knapsack instance}
      \label{fig:distr-Deb-easy-kp}
   \end{subfigure}
   \begin{subfigure}{0.3\textwidth}
      \centering
      \includegraphics[width=\textwidth]{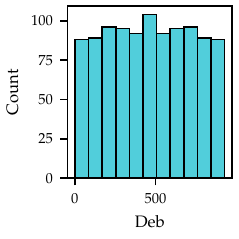}
      \caption{Hard knapsack instance}
      \label{fig:distr-Deb-hard-kp}
   \end{subfigure}
   \caption{Histograms of Deb's feasibility rule ranks.}
\end{figure}

\begin{figure}[ht!]
   \centering
   \begin{subfigure}{0.3\textwidth}
      \centering
      \includegraphics{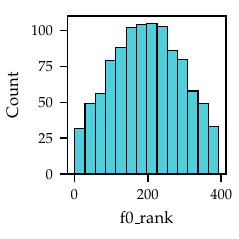}
      \caption{Easy knapsack instance}
      \label{fig:distr-f-rank-easy-kp}
   \end{subfigure}
   \begin{subfigure}{0.3\textwidth}
      \centering
      \includegraphics{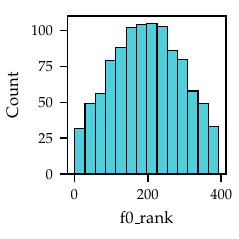}
      \caption{Hard knapsack instance}
      \label{fig:distr-f-rank-hard-kp}
   \end{subfigure}
   \caption{Histograms of objective ranks.}
\end{figure}

\paragraph{Deb's Feasibility Rule Ranks Distribution (\texttt{distr\_Deb}):}
\citet{deb2000efficient}'s feasibility rule is used to rank solutions based on three criteria: (1) feasible solutions are better than infeasible solutions, (2) among two feasible solutions, the one with a better objective value is better, and (3) among infeasible solutions, solutions with lower violation values are better.
The \texttt{distr\_Deb} feature describes the distribution of Deb's ranks by producing a histogram and calculating descriptive statistics.
The \texttt{distr\_Deb} feature ranks solutions of the easy and hard knapsack instances as in the visual output in \Cref{fig:distr-Deb-easy-kp,fig:distr-Deb-hard-kp} and the numerical outputs in \Cref{tab:distr-Deb}.
The Deb's ranks for the easy knapsack are normally distributed because they follow the distribution of the objective ranks (\Cref{fig:distr-f-rank-easy-kp}): the almost constant distribution of violation ranks (\Cref{fig:distr-v-easy-kp}) has little impact on the combined ranking.
By contrast, the Deb's ranks for the hard knapsack instance are uniformly distributed, whereas the objective (\Cref{fig:distr-f-rank-hard-kp}) and violation ranks (\Cref{fig:distr-v-hard-kp}) are normally distributed. Combining objective and constraint ranking criteria results in more distinct ranks, hence the uniform distribution.

Deb's ranks are used in other features such as \texttt{corr\_ranks}, \texttt{RDC}, \texttt{PDC}, \texttt{disp\_best}, and \texttt{NRC}. This is because Deb's ranks provide an aggregate view of all objective and violation ranks in the landscape, allowing comparisons of objective and violation value(s) with an overall aggregate.

\begin{table}[ht!]
   \centering
   \caption{Output of \texttt{distr\_Deb} Descriptive statistics of Deb's feasibility rule ranks for the easy and hard knapsack instances.}
   \label{tab:distr-Deb}
   \begin{tabular}{l|rr}
      \toprule
      \textbf{Statistic} & \textbf{Easy knapsack instance} & \textbf{Hard knapsack instance}\\
      \midrule
      Minimum & 1 & 1\\
      Maximum & 463 & 928\\
      Mean & 205.0576 & 464.5\\
      Median & 197.0 & 464.5\\
      Q1 & 126.0 & 238.0\\
      Q3 & 274.0 & 691.0\\
      Sd & 107.1946 & 263.4005\\
      Skewness & 0.3704 & 0.0\\
      Kurtosis & -0.4335 & -1.1692\\
      \bottomrule
   \end{tabular}
\end{table}

\subsubsection{Distance-based Features}\label{subsubsec:dist-feat}
The illustrative use cases for the distance-based features are discussed in this section.

\begin{figure}[ht!]
   \centering
   \begin{subfigure}{0.45\textwidth}
      \centering
      \includegraphics{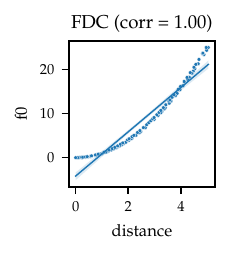}
      \caption{Sphere problem}
      \label{fig:fdc-sphere}
   \end{subfigure}
   \begin{subfigure}{0.45\textwidth}
      \centering
      \includegraphics{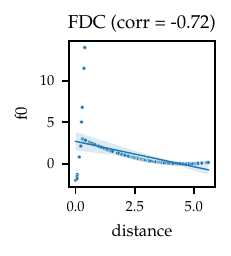}
      \caption{Deceptive problem}
      \label{fig:fdc-deceptive}
   \end{subfigure}
   \caption{\texttt{FDC} visual output for the constrained sphere and constrained deceptive problems.}
\end{figure}

\paragraph{Fitness Distance Correlation (\texttt{FDC}):} The \texttt{FDC} feature is implemented following~\cite{jones1995fitness}, but using the best solution in the input data in place of the known global optimum. It is applied to the sphere problem and the deceptive problem to characterise their level of deception when searching locally. \Cref{fig:fdc-sphere,fig:fdc-deceptive} show the visual output of the \texttt{FDC} feature. According to \Cref{fig:fdc-sphere}, for the sphere problem, there is an almost perfect correlation between the objective values and distance to the nearest best solution. This means that the search space is not deceptive, in that a search in the solution space will likely arrive at the global minimum since solutions become fitter and fitter as the search gets closer to the best solution. Conversely, \Cref{fig:fdc-deceptive} shows a high negative correlation of -0.72 between distance to the nearest best solution and objective values. Therefore, nearer the best solution, solutions tend to deteriorate in fitness, signalling deceptive information for search. In this case, the \texttt{FDC} indicates that the problem is indeed deceptive.

\begin{figure}[ht!]
   \begin{subfigure}{0.3\textwidth}
      \includegraphics{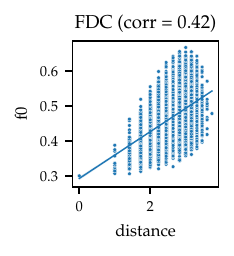}
      \caption{$k=1$}
      \label{fig:fdc-nk-k1}
   \end{subfigure}
   \hfill
   \begin{subfigure}{0.3\textwidth}
      \includegraphics{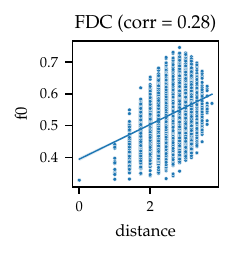}
      \caption{$k=2$}
      \label{fig:fdc-nk-k2}
   \end{subfigure}
   \hfill
   \begin{subfigure}{0.3\textwidth}
      \includegraphics{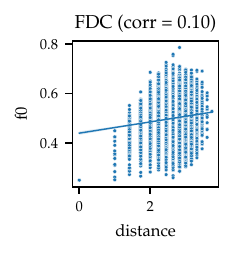}
      \caption{$k=4$}
      \label{fig:fdc-nk-k4}
   \end{subfigure}
   \caption{\texttt{FDC} visual output for NK landscape instances.}
\end{figure}

The \texttt{FDC} feature is further used to characterise the three NK landscape instances with increasing difficulty as $k$ increases. For values of $k=1$, 2, 4, the numerical output is 0.42, 0.28, 0.10, respectively, indicating that increasing values of $k$ reduce the likelihood of reaching the global optimum, as the relationship between a solution's fitness and its distance to the global solution weakens with increasing values of $k$. \Cref{fig:fdc-nk-k1,fig:fdc-nk-k2,fig:fdc-nk-k4} show the corresponding visual output.

\begin{figure}[htb!]
\centering
   \begin{subfigure}{0.3\textwidth}
      \includegraphics{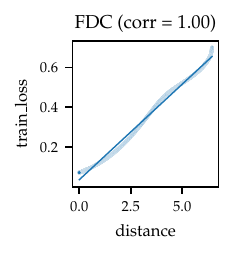}
      \caption{SGD trajectory}
      \label{fig:fdc-xor-nn-trained}
   \end{subfigure}
   \begin{subfigure}{0.3\textwidth}
      \includegraphics{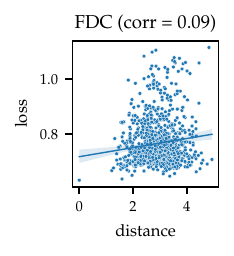}
      \caption{Random sampling}
      \label{fig:fdc-xor-nn-sampled}
   \end{subfigure}
   \caption{Output of \texttt{FDC} for scenarios simulating the XOR gate using a neural network: (\subref{fig:fdc-xor-nn-trained}) corresponds to the SGD trajectory in parameter space, and (\subref{fig:fdc-xor-nn-sampled}) corresponds to the scenario where neural network parameters are randomly sampled.}
\end{figure}

\Cref{fig:fdc-xor-nn-trained,fig:fdc-xor-nn-sampled} show the output of \texttt{FDC} for the scenarios simulating the XOR gate using a neural network.
The SGD trajectory trace has an almost perfect fitness-distance correlation, indicating that for the XOR gate problem, SGD effectively minimises binary cross-entropy loss and does not get stuck in local minima.
For the second scenario, which involves randomly sampled parameters, a fitness distance correlation of 0.09 shows that the distance between any two solutions is not correlated with the distance between the two solutions in the objective space.
Therefore, finding good solutions at random is not trivial.
Good solutions are readily found when there is gradient information, as is the case for optimisation using SGD.

\begin{figure}[ht!]
   \centering
   \includegraphics{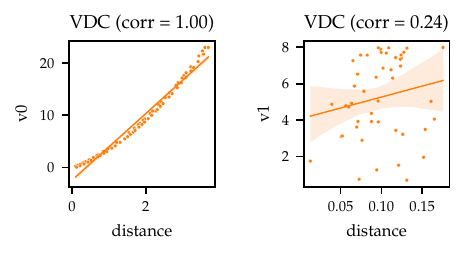}
   \caption{Visual output of the \texttt{VDC} feature for each constraint of the constrained sphere problem, where $v0$ and $v1$ are Constraints~(\ref{eq:sphere-constraint}) and (\ref{eq:sine-constraint}).}
   \label{fig:vdc}
\end{figure}

\paragraph{Violation Distance Correlation (\texttt{VDC}):} This feature is implemented by extending the \texttt{FDC} feature~\citep{jones1995fitness} to violation values.
The \texttt{VDC} feature is evaluated on the constrained sphere problem, resulting in the visual output shown in \Cref{fig:vdc}. The violation values of the infeasible solutions of Constraint~(\ref{eq:sphere-constraint}) are almost perfectly correlated with the distance to the nearest feasible solutions. This implies that it is easy to cross to a feasible region from an infeasible region with respect to Constraint~(\ref{eq:sphere-constraint}). Conversely, the low Spearman's correlation coefficient (0.24) of Constraint~(\ref{eq:sine-constraint}) indicates a lower likelihood of crossing from an infeasible region to a feasible region.

\begin{figure}[ht!]
   \includegraphics{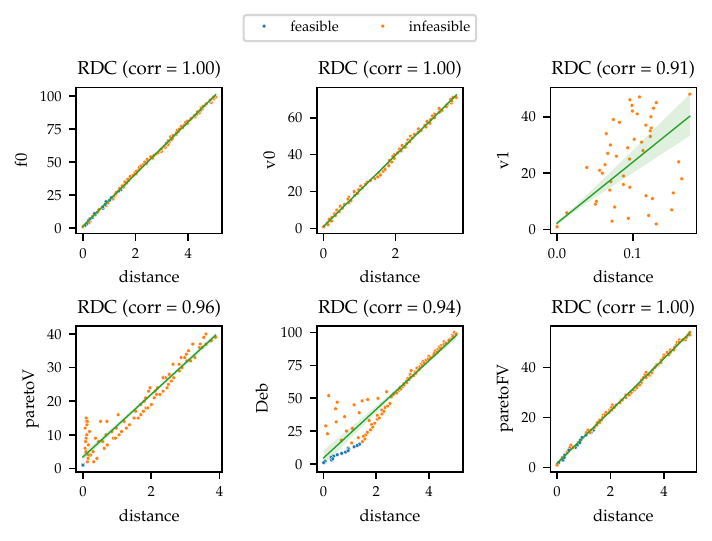}
   \caption{Visual output of the \texttt{RDC} feature for each rank computed for the constrained sphere problem.}
   \label{fig:rdc}
\end{figure}

\paragraph{Rank Distance Correlation (\texttt{RDC}):} \texttt{RDC} is implemented by extending \texttt{FDC}~\citep{jones1995fitness} to ranks; see also \citet{ochoa_ppsn2024}.
The \texttt{RDC} feature is used to analyse the ranks of the constrained sphere problem. \Cref{fig:rdc} shows the visual output, while the corresponding numerical outputs (rounded to two decimal places) are as follows: 1.00 for the objective, 1.00 for Constraint~(\ref{eq:sphere-constraint}), 0.91 for Constraint~(\ref{eq:sine-constraint}), 0.96 for $paretoV$, 0.94 for Deb's rank, and 1.00 for $paretoFV$. The objective rank, the Constraint~(\ref{eq:sphere-constraint}) violation values rank, the Pareto ranks ($paretoV$, $paretoFV$), and Deb's rank are found to have very high Spearman's correlation coefficients ranging from 0.94 to 1.00. The rank for Constraint~(\ref{eq:sine-constraint}) exhibits a lower Spearman's correlation coefficient of 0.91, which remains a high value. This means that local search easily finds improving solutions with respect to fitness and constraint violation, though less easily with respect to Constraint~(\ref{eq:sine-constraint}). 
This result is consistent with the output of the \texttt{corr} and \texttt{corr\_ranks} features.

\begin{figure}[ht!]
    \centering
    \includegraphics{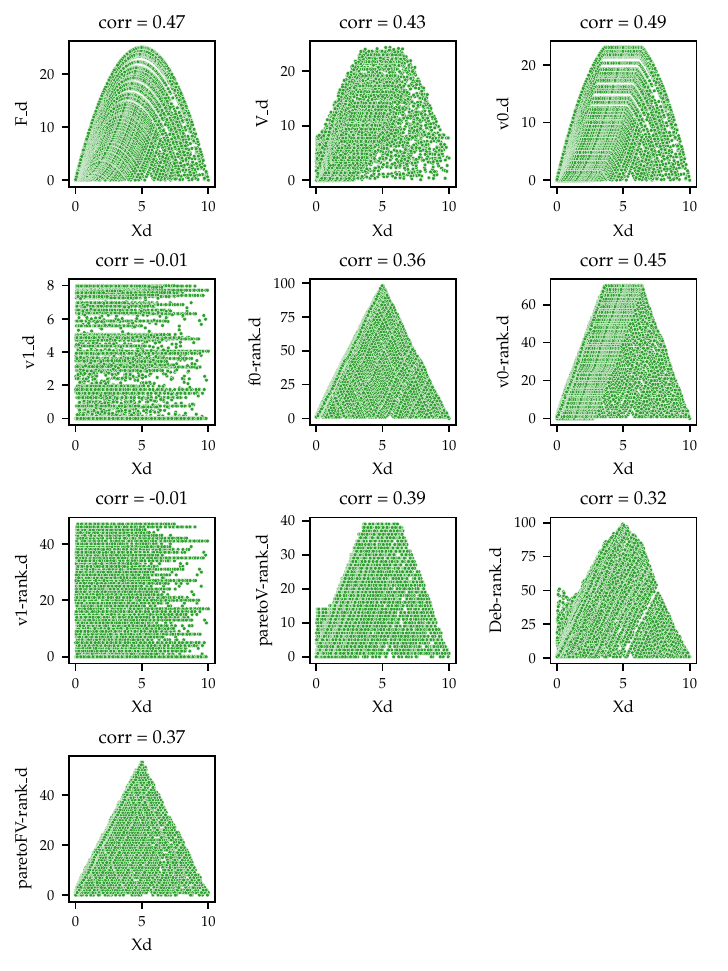}
    \caption{Visual output of the \texttt{PDC} feature for the constrained sphere problem.}
    \label{fig:pdc-sphere}
\end{figure}

\begin{figure}[ht!]
    \centering
    \includegraphics{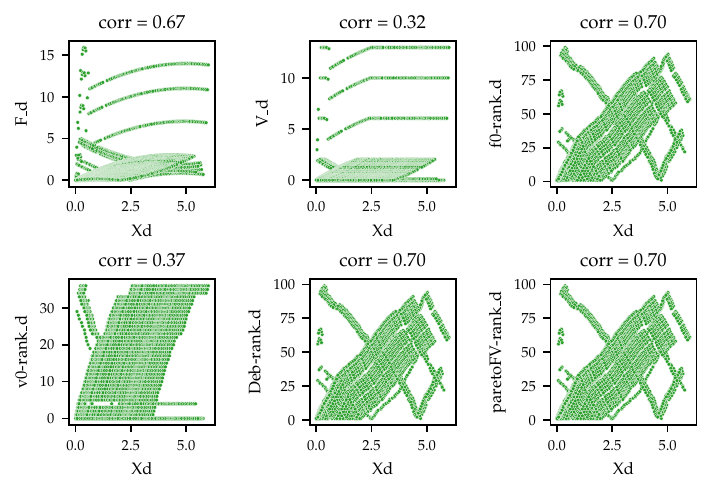}
    \caption{Visual output of the \texttt{PDC} feature for the constrained deceptive problem.}
    \label{fig:pdc-deceptive}
\end{figure}

\begin{figure}[ht!]
    \centering
        \includegraphics{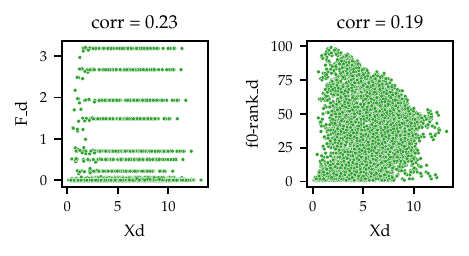}
    \caption{Visual output of the \texttt{PDC} feature for the Easom problem.}
    \label{fig:pdc-easom}
\end{figure}

\paragraph{Pairwise Distance Correlation (\texttt{PDC}):} \texttt{PDC} produces scatter plots of pairwise distance in the variable space against pairwise distance in the objective space (all objectives), violation space (all violations), and against each objective, violation, and rank. The corresponding numerical outputs are the Spearman's correlation coefficients between each set of pairwise distances.
The \texttt{PDC} feature is applied to the constrained sphere problem, constrained deceptive problem, and Easom function problem, resulting in the visual output shown in \Cref{fig:pdc-sphere,fig:pdc-deceptive,fig:pdc-easom}, respectively, and the corresponding numerical outputs in \Cref{tab:pdc-sphere,tab:pdc-deceptive}. The scatter plots evoke the shapes of the underlying functions transformed to two dimensions, showing all three problems to be substantially different from each other. The objective of the constrained sphere problem has a weak correlation value of 0.47, which is expected, as the sphere function is not monotonically increasing. Monotonically increasing functions will show a high correlation since, as pairwise distance on the x-axis increases, pairwise distance on the y-axis also increases. The Easom function problem has the lowest correlation values: objective space \texttt{PDC} of 0.23 and objective rank \texttt{PDC} of 0.19. This is explained by its neutrality property, which causes the relationship between pairwise distance in the variable space and in the objective space to be the weakest among the three.

\begin{table}[ht!]
   \centering
   \caption{Numerical output of \texttt{PDC} for the constrained sphere problem.}
   \label{tab:pdc-sphere}
   \begin{tabular}{l|r}
      \toprule
      \textbf{Value/Rank space} & \textbf{Sphere}\\
      \midrule
      Objective space($F$) & 0.4393 \\
      Violation space ($V$) & 0.4116 \\
      \Cref{eq:sphere-constraint} constraint& 0.4566 \\
      \Cref{eq:sine-constraint} constraint& 0.0098 \\
      Objective rank & 0.3433 \\
      Violation rank & 0.4233 \\
      Rank of \Cref{eq:sphere-constraint} constraint& 0.4233 \\
      Rank of \Cref{eq:sine-constraint} constraint& 0.0155 \\
      $paretoV$ rank & 0.3588 \\
      $Deb$ rank & 0.3194 \\
      $paretoFV$ rank & 0.2929 \\
      \bottomrule
   \end{tabular}
\end{table}

\begin{table}[ht!]
   \centering
   \caption{Numerical output of \texttt{PDC} for the constrained deceptive problem.}
   \label{tab:pdc-deceptive}
   \begin{tabular}{l|r}
      \toprule
      \textbf{Value/Rank space} & \textbf{Deceptive}\\
      \midrule
      Objective space ($F$) & 0.6251\\
      Violation space ($V$) & 0.3155\\
      Objective rank & 0.6413\\
      Violation rank & 0.3274\\
      $Deb$ rank & 0.6413\\
      $paretoFV$ & 0.6413\\
      \bottomrule
   \end{tabular}
\end{table}

\begin{figure}[ht!]
   \centering
   \includegraphics{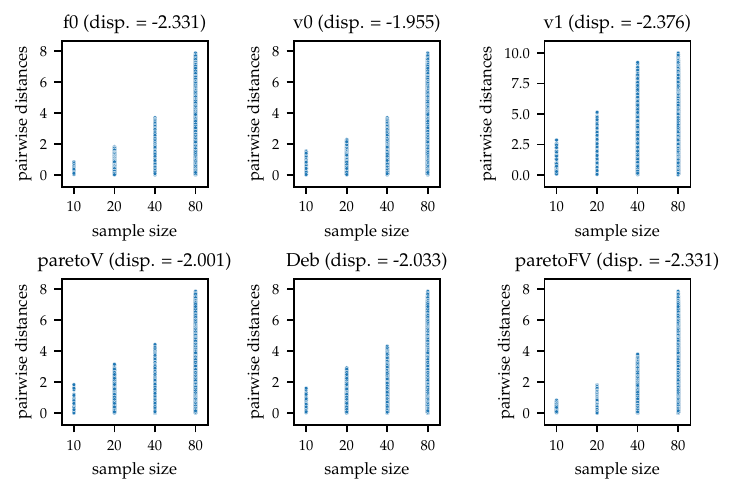}
   \caption{Visual output of \texttt{disp\_best} for the constrained sphere problem; \emph{disp.} is an abbreviation for dispersion.}
   \label{fig:disp-best-sphere}
\end{figure}

\begin{figure}[ht!]
   \centering
   \includegraphics{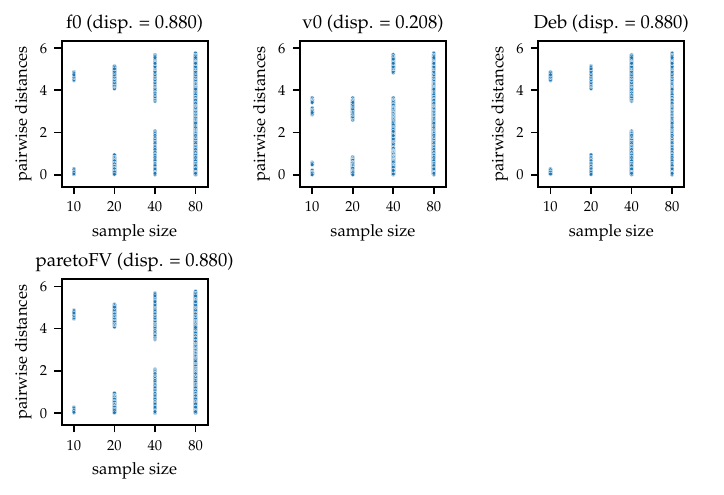}
   \caption{Visual output of \texttt{disp\_best} for the constrained deceptive problem; \emph{disp.} is an abbreviation for dispersion}
   \label{fig:disp-best-deceptive}
\end{figure}

\paragraph{Dispersion of the Best Solutions (\texttt{disp\_best}):} The implementation of the \texttt{disp\_best} feature is based on the dispersion metric proposed by~\cite{lunacek2006dispersion}.
The \texttt{disp\_best} feature is used to characterise the constrained sphere and constrained deceptive problems with regard to multimodality and global topology. \Cref{fig:disp-best-sphere,fig:disp-best-deceptive} show the visual output for the constrained sphere and constrained deceptive problems, respectively. Considering the objective ranks, the constrained sphere problem's objective has a negative dispersion metric value of -2.331, which indicates the presence of a unimodal global structure, as expected. The constrained deceptive problem, on the other hand, has a positive dispersion metric value of 0.88 for the objective, indicating the presence of funnels, as can be concluded from the line plot of its objective values in \Cref{fig:deceptive}. The scatter plots visually express the dispersion, where for the constrained sphere problem, the spread of pairwise distances grows as the sample size $n$ of the best solution is increased. Conversely, for the constrained deceptive problem, the gaps between the clusters of points for lower sample sizes show that indeed the best solutions are much further apart, accounting for the presence of two funnels.

\subsubsection{Neighbourhood-based Features}\label{subsubsec:neighbourhood-feat}
\begin{figure}[ht!]
    \centering
    \begin{subfigure}{0.3\textwidth}
        \includegraphics{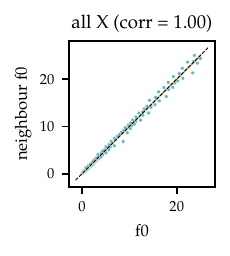}
        \caption{Constrained sphere (all solutions)}
        \label{fig:nfc-sphere-all}  
    \end{subfigure}
    \begin{subfigure}{0.3\textwidth}
        \includegraphics{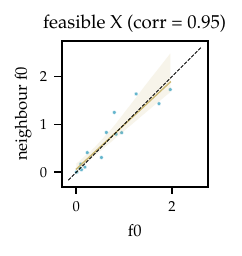}
        \caption{Constrained sphere (feasible only)}
        \label{fig:nfc-sphere-feasible}  
    \end{subfigure}
    \begin{subfigure}{0.3\textwidth}
        \includegraphics{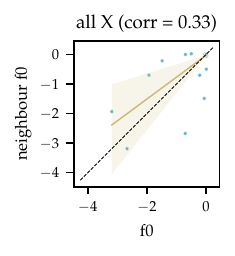}
        \caption{Easom function (all solutions)}
        \label{fig:nfc-easom}  
    \end{subfigure}
    \caption{Visual output \texttt{NFC}.}
\end{figure}

The use cases for the neighbourhood-based features are presented below:

\paragraph{Neighbouring Solutions' Objective (Fitness) Values Correlation (\texttt{NFC}):} \texttt{NFC} is implemented following the fitness cloud technique proposed by~\cite{verel2003bottlenecks}.
The \texttt{NFC} feature is applied to the constrained sphere problem and the Easom function problem to distinguish them in terms of the evolvability of their landscapes. \Cref{fig:nfc-sphere-all,fig:nfc-sphere-feasible,fig:nfc-easom} show the visual output for all the solutions of the constrained sphere problem, for the feasible solutions of the constrained sphere problem, and for all the solutions of the Easom function problem (unconstrained), respectively. The constrained sphere problem is found to have high evolvability, evidenced by a high \texttt{NFC} value of 1 and 0.95 for all solutions and for feasible solutions only. This is expected, since the sphere function is one of the simplest continuous optimisation problems. The Easom function problem, on the other hand, has a weaker \texttt{NFC} value of 0.33. This is because the Easom function's landscape is characterised by neutrality; therefore, there is a lower likelihood of moving to a solution of better fitness in the search space.

\begin{figure}[ht!]
    \centering
    \includegraphics{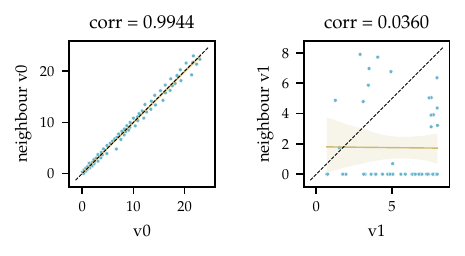}
    \caption{Visual output of the \texttt{NVC} feature for the constrained sphere problem.}
    \label{fig:nvc}
\end{figure}

\paragraph{Neighbouring Solutions' Violation Values Correlation (\texttt{NVC}):} \texttt{NVC} is implemented by extending the fitness cloud technique~\citep{verel2003bottlenecks} to the violation values.
The \texttt{NVC} feature is used to characterise the constraints of the constrained sphere problem, resulting in the visual output shown in \Cref{fig:nvc}. Constraint~(\ref{eq:sphere-constraint}) ($v0$), being a simple sphere function, shows an almost perfect correlation of 0.9944. This indicates that there is a high searchability with respect to Constraint~(\ref{eq:sphere-constraint}): it is easy to reach solutions of improving feasibility. Conversely, Constraint~(\ref{eq:sine-constraint}) ($v1$) shows a weak correlation of $0.036$ since, due to its cyclic nature, there is no strong relationship between the violation values of neighbouring solutions. It is thus less likely for the search to continuously traverse solutions of improving feasibility. Therefore, there is low searchability with respect to Constraint~(\ref{eq:sine-constraint}).

\begin{figure}[ht!]
    \centering
    \includegraphics{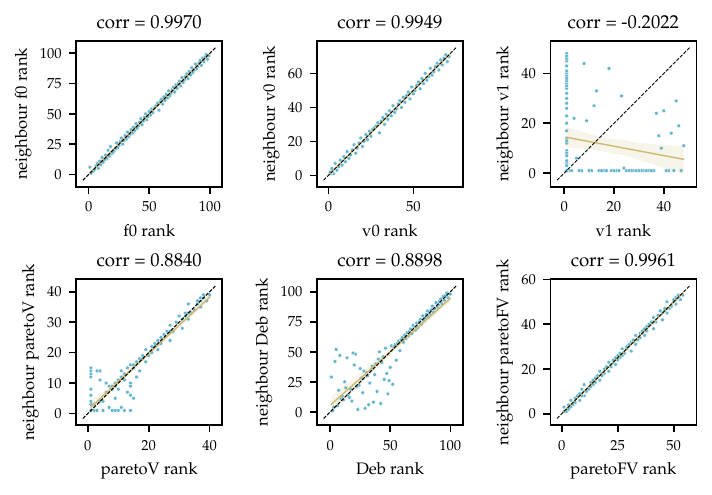}
    \caption{Visual output of \texttt{NRC} for the constrained sphere problem.}
    \label{fig:nrc-sphere}
\end{figure}

\begin{figure}[ht!]
   \centering
   \includegraphics{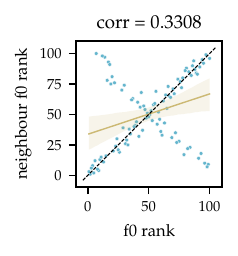}
   \caption{Visual output of \texttt{NRC} for the Easom function problem.}
   \label{fig:nrc-easom}
\end{figure}

\paragraph{Neighbouring Solutions' Ranks Correlation (\texttt{NRC}):} \texttt{NRC} is implemented by extending the fitness cloud technique~\citep{verel2003bottlenecks} to ranks, such that fitness values are substituted with ranks.
The \texttt{NRC} feature is used to comparatively analyse the constained sphere problem and the Easom function problem. \Cref{fig:nrc-sphere,fig:nrc-easom} show the visual output of the \texttt{NRC} feature for the constrained sphere problem and the Easom function problem. The corresponding numerical output for the constrained sphere problem is as follows: 0.9970 for the objective ranks, 0.9949 for the ranks of Constraint~(\ref{eq:sphere-constraint}), -0.2022 for the ranks of Constraint~(\ref{eq:sine-constraint}), 0.8840 for $paretoV$ ranks, 0.8898 for Deb's ranks, and 0.9961 for $paretoFV$ ranks. 
The Easom function is unconstrained and has only one objective, hence it only has one rank. The objective of the constrained sphere problem is found to have high searchability given its almost perfect \texttt{NRC} coefficient, while the Easom function, due to its neutrality, is much less searchable with a \texttt{NRC} coefficient of 0.3308. Further, all the ranks of the sphere function have a high \texttt{NRC} except the ranks of Constraint~(\ref{eq:sine-constraint}), which, due to the cyclic nature of the sine function, is not highly searchable.

\begin{figure}[ht!]
\centering
   \begin{subfigure}{0.4\textwidth}
      \includegraphics{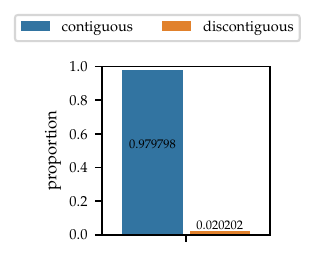}
      \caption{Constraint~(\ref{eq:sphere-constraint})}
      \label{fig:ncf-v0}  
   \end{subfigure}
    \begin{subfigure}{0.4\textwidth}
      \includegraphics{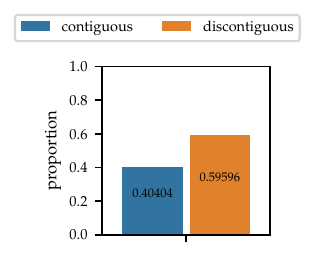}
      \caption{Constraint~(\ref{eq:sine-constraint})}
      \label{fig:ncf-v1}  
   \end{subfigure}
   \begin{subfigure}{0.4\textwidth}
      \includegraphics{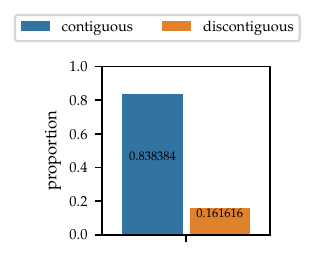}
      \caption{Constraints (\ref{eq:sphere-constraint}) and (\ref{eq:sine-constraint}) combined}
      \label{fig:ncf-all-v}  
   \end{subfigure}
   \caption{Bar chart showing output of \texttt{NCF} for the constrained sphere problem.}
\end{figure}

\paragraph{Neighbouring Change in Feasibility (\texttt{N$\Delta$Feas}):} \verb|NCF| uses a neighbourhood (\verb|N|) file to compute the proportion of transitions from one solution to another that result in a change in feasibility and the proportion of transitions that do not result in a change in feasibility.
\Cref{fig:ncf-v0,fig:ncf-v1} show the visual output of the \verb|NCF| feature. The numerical outputs, which are the proportions, are annotated onto the bars of the bar chart. For Constraint~(\ref{eq:sphere-constraint}), the proportion of contiguous feasibility (0.9798) is much larger than the proportion of discontiguous feasibility (0.0202). This is expected since, as shown in \Cref{fig:sphere-v0}, the feasible region is contiguous and is sandwiched between two contiguous infeasible regions. By contrast, Constraint~(\ref{eq:sine-constraint}) has the proportion of discontiguous feasibility (0.596) higher than the proportion of contiguous feasibility (0.404) as expected, since the sine function is cyclic, causing a higher likelihood of transitions ending up on a neighbour of dissimilar feasibility compared to its predecessor.
When the two constraints are considered, the proportion of contiguous feasibility (0.8383) is still much larger than the proportion of discontiguous feasibility (0.1616) as shown in \Cref{fig:ncf-all-v}.

\section{Conclusion}\label{sec:conclusion}
This paper proposed a flexible and generic approach for e\textbf{x}plainable \textbf{l}andscape \textbf{a}nalysis (XLA) and illustrated the utility of its associated toolbox, \texttt{pyXla}.
The proposed approach is generic, since it allows for analysis on any problem domain: continuous or combinatorial, constrained or unconstrained, single-objective or multi-objective alike.
This generality and flexibility are achieved by the separation of input representation and sampling from actual analysis.
Explainability is achieved by the implementation of intuitively interpretable LA techniques which produce both visual and numerical outputs. The proposed framework, empowered by \texttt{pyXla}, anticipates a shift toward XLA. Explainability and interpretability are key aims of LA, aims that are overshadowed by black-box approaches 
relying on automated machine learning pipelines to classify uninterpretable features.
The \texttt{pyXla} toolbox exposes a flexible and generic input representation interface and conveniently composes the 16 explainable LA features and five sampling algorithms, thus simplifying explainable landscape analysis.

In the proposed approach, inputs are either provided in designated data files or by specifying functions to generate the inputs. The input data is: \texttt{F} for objective values, \texttt{X} for solutions, \texttt{V} for violation values, \texttt{D} for pairwise distances between solutions, \texttt{N} for neighbourhood relationships between solutions, and \texttt{I} for any other additional input. 
Additionally, five sampling techniques are provided: uniform random, Latin hypercube, random walk, adaptive walk, and Hilbert curve sampling.
Random uniform sampling applies to both combinatorial and continuous domains, while the rest are for continuous domains in the current \texttt{pyXla} version.
A total of 16 explainable landscape analysis features grouped into 4 categories are available, comprising the statistical features: distribution of objective values (\texttt{distr\_F}), distribution of violation values (\texttt{distr\_V}), correlation of values (\texttt{corr}), correlation of ranks (\texttt{corr\_ranks}), and variable importance (\texttt{X\_imp}); the rank-based features: distribution of Pareto ranks (\texttt{distr\_Par}) and distribution of Deb's ranks (\texttt{distr\_Deb});  the distance-based features: fitness distance correlation (\texttt{FDC}), violation distance correlation (\texttt{VDC}), rank distance correlation (\texttt{RDC}), pairwise distance correlation (\texttt{PDC}), and dispersion of the best solutions (\texttt{disp\_best}); and the neighbourhood features: neighbouring solutions' objective values correlation (\texttt{NFC}), neighbouring solutions' violation value of correlation (\texttt{NVC}), neighbouring solutions' ranks correlation (\texttt{NRC}), and neighbouring change in feasibility (\texttt{NCF}).

Future work lies in using \texttt{pyXla} to analyse an entire problem suite, real-world problems, and adding more XLA features and sampling techniques. The developed toolbox is not exhaustive in its collection of XLA features. We plan to include the following XLA features in future versions: local optima networks~\citep{ochoa2014}, ruggedness measure based on entropy~\citep{malan2009quantifying}, neutrality measures $\mathcal{M}_1$ and $\mathcal{M}_2$~\citep{van2017characterising},
multi-objective measures based on dominance and indicators~\citep{liefooghe_tevc2020},
a non-linear variable importance measure based on influence of mutation of a variable on fitness~\citep{lepretre2019fitness},
as well as features that capture variable interactions.



\section*{Acknowledgements}
KM would like to thank LISIC, University of Littoral Côte d'Opale for financially supporting her research visits in October 2024, which led to this work.
This research was supported by the University Capacity Development Programme (UCDP) of the Department of Higher Education and Training (DHET) of South Africa.
This research was carried out in part within the Randomised Optimisation Algorithms Research Network (ROAR-NET, CA22137), supported by COST (European Cooperation in Science and Technology).

\bibliographystyle{apalike}
\bibliography{ecjsample}

\end{document}